\documentclass{article}

\PassOptionsToPackage{numbers,compress}{natbib}

\usepackage[preprint]{neurips_2026}

\usepackage[utf8]{inputenc} 
\usepackage[T1]{fontenc}    
\usepackage{hyperref}       
\usepackage{url}            
\usepackage{booktabs}       
\usepackage{amsfonts}       
\usepackage{nicefrac}       
\usepackage{microtype}      
\usepackage{xcolor}         
\usepackage{graphicx}
\usepackage{enumitem}
\usepackage{amsmath} 
\usepackage{multicol}
\usepackage{multirow}
\usepackage{caption}
\usepackage[capitalise]{cleveref}
\usepackage[font=small]{caption} 
\crefname{section}{Appendix}{Appendices}

\title{BrainExplore: Large-Scale Discovery of Interpretable Visual Representations in the Human Brain}

\author{%
  \normalfont
  Navve Wasserman$^{1,*}$ \quad
  Matias Cosarinsky$^{1,*}$ \quad
  Yuval Golbari$^{1}$ \quad
  Aude Oliva$^{2}$ \\
  Antonio Torralba$^{2}$ \quad
  Tamar Rott Shaham$^{2}$ \quad
  Michal Irani$^{1}$ \\[6pt]
  {\small $^{1}$Weizmann Institute of Science} \quad
  {\small $^{2}$Massachusetts Institute of Technology} \quad
  {\small $^{*}$Equal contribution}
}

\begin{document}

\maketitle

\vspace{-0.3cm}
\begin{abstract}
    Understanding how the human brain represents visual concepts, and in which brain regions these representations are encoded, remains a long-standing challenge. Decades of work have advanced our understanding of visual representations, yet brain signals remain large and complex, and the space of possible visual concepts is vast. As a result, most studies remain small-scale, rely on manual inspection, focus on specific regions and concepts, and rarely include systematic validation.
    We present a large-scale, automated framework for discovering and explaining visual representations across the human cortex. Our method comprises two main stages. First, we \emph{discover} candidate interpretable patterns in fMRI activity through unsupervised, data-driven decomposition methods. Next, we \emph{explain} each pattern by identifying the set of natural images that most strongly elicit it and generating a natural-language description of their shared visual meaning. To scale this process, we introduce an automated pipeline that tests multiple candidate explanations, assigns reliability scores, and selects the best description for each voxel pattern. Our framework reveals thousands of interpretable patterns spanning many distinct visual concepts, including fine-grained representations previously unreported. For a demo, models and labeled data, see our \href{https://navvewas.github.io/BrainExplore/}{project page}.
\end{abstract}

\section{Introduction}
\label{sec:intro}
\vspace{-0.15cm}

Understanding how the human brain represents visual information is a long-standing challenge. The visual cortex encodes a rich hierarchy of features that support object recognition, scene understanding, and visual reasoning. Yet, the structure of these representations and their underlying organization in the brain remain largely unknown. 
Functional Magnetic Resonance Imaging (fMRI) has become a dominant tool for studying how the human brain processes visual information, providing a non-invasive window into cortical activity while participants view natural images~\cite{allen2022massive,Horikawa2017GenericDecoding}. It measures brain activity 
across the brain, parceled into tiny volume elements
 (``voxels''). Over the past decades, fMRI researchers have sought to interpret these complex voxel-level patterns to uncover the structure of visual representations in the brain. 
Early work targeted retinotopic maps and low-level features~\cite{Engel1997CerebCortex,Sereno1995Science,DeYoe1996PNAS,Kamitani2005NatNeurosci,Tootell1995Nature}, while later studies examined higher-level semantics using category-selective analyses~\cite{weiner2011not,downing2001cortical,Adamson18,Jain23,kanwisher2006fusiform}, data-driven fMRI decompositions~\cite{ecker2007detecting,Huth12,van2024core,zhao2024separate,Khosla22} and image–brain models~\cite{ccukur2013functional,hwang2025silico,braindiffusion,neurogen}.

However, fMRI signals are high-dimensional, comprising tens of thousands of brain voxels per subject that reflect a vast range of possible visual concepts, while available image-to-fMRI datasets remain relatively small. As a result, current studies often focus on specific properties (e.g., faces, places, food), or on particular brain regions, such as face-selective (FFA) or place-selective (PPA). Moreover, analyses typically rely on manual inspection, making it difficult to scale interpretations across many fMRI patterns and visual concepts. 
These limitations highlight the need for scalable, automated tools capable of discovering and explaining visual representations across the entire cortex.

Recent advances in interpretability of artificial neural networks, such as Language and Vision models, have shown that high-level concepts can be \textit{automatically} discovered from internal activations~\cite{bau2017network,hernandez2021natural,gandelsman2023interpreting}. These results suggest that similar data-driven approaches might offer new insight into the representations that emerge in the brain. Inspired by this, we introduce \emph{BrainExplore}, an unsupervised, data-driven framework for large-scale discovery and evaluation of visual concept representations across the visual cortex. BrainExplore automatically discovers interpretable patterns of brain activity and explains them in a natural language. 

\begin{figure*}[t]
  \centering
\includegraphics[width=\textwidth]{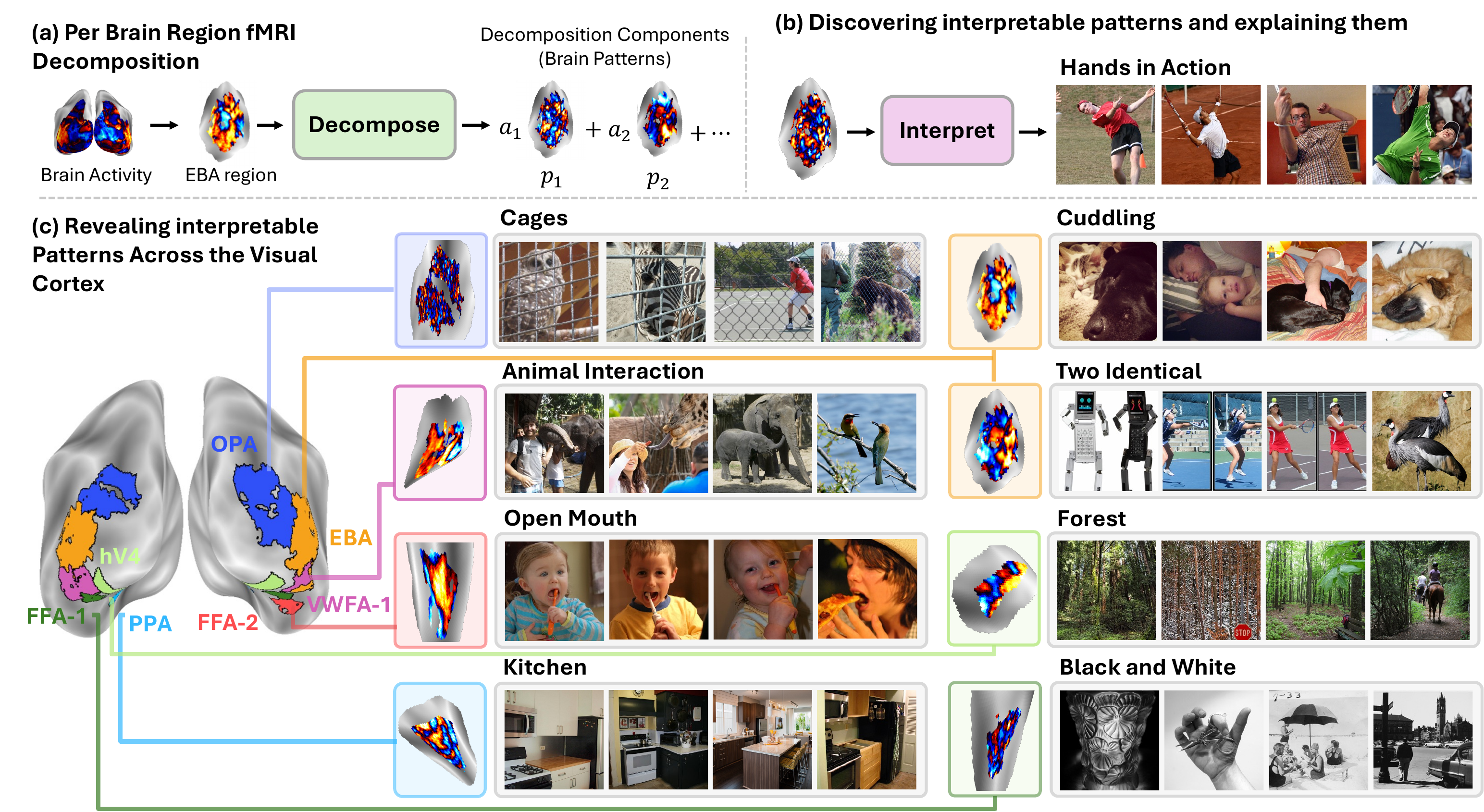}
  \vspace{-0.45cm}
\caption{\textbf{BrainExplore: Discovering Interpretable Visual Representations in the Human Brain.}
\small \textbf{(a)} Per–region fMRI decomposition learns component patterns such that any response is approximated as a linear combination of these patterns. 
\textbf{(b)} For each pattern, we retrieve its top-\(N\) activating fMRI responses and corresponding images, then automatically select the best matching concept and assign an alignment score. We then use these scores to identify the most interpretable patterns and their explanations.
\textbf{(c)} Examples of discovered interpretable patterns across regions, showing pattern activations projected onto cortex, top activating images, and their textual explanations.}
  \vspace{-0.54cm}
  \label{fig:teaser}
\end{figure*}

Our pipeline comprises two main steps: \textbf{(i)} deriving candidate interpretable fMRI patterns via per–brain-region decomposition (\cref{fig:teaser}a); and \textbf{(ii)} discovering interpretable patterns and explaining them (\cref{fig:teaser}b).
 The first step performs per brain region decomposition of fMRI activity to extract components that may correspond to meaningful visual representations. Conceptually, each decomposition method learns a set of fMRI patterns (components) such that any fMRI response in a given region of interest (``ROI'') can be approximated as a linear combination of these patterns. For each component, we retrieve the top-\(N\) fMRI responses with the highest coefficient for that component and collect their corresponding images. This yields a set of images per component that \emph{visualize} the visual concept that most strongly activates it.
We then discover interpretable patterns by automatically identifying, for each component, the concept that best aligns with it and assigning an alignment score. This allows us to surface the most interpretable components together with their explanations.

Using \emph{BrainExplore}, we discover thousands of interpretable patterns throughout the brain. Since the method automatically evaluates the quality of each explanation, it enables integration of findings across different decomposition methods rather than relying on a single one. We exploit established techniques (PCA~\cite{hotelling1933analysis}, NMF~\cite{lee1999learning}, ICA~\cite{comon1994independent}) as well as Sparse Autoencoders~\cite{cunningham2023sparseautoencodershighlyinterpretable} (SAEs), a method widely used to interpret artificial neural networks via projection into high-dimensional space and sparsity constraints. SAE yields a large number of interpretable patterns and reveals visual representations that are complementary to other methods, including representations not captured elsewhere.  
Moreover, to overcome limited data, we leverage an image-to-fMRI prediction model~\cite{beliy2024wisdom} that synthesizes brain responses for unseen images. This allows us to expand the dataset from approximately 10k images with measured fMRI to over 120k images with measured or predicted responses. This augmentation improves decomposition quality and increases the diversity of discovered representations.

Altogether, \emph{BrainExplore} reveals a large set of nuanced representations across the brain. These include interpretable patterns selective for object identity; for people, body parts, and pose (e.g., open mouths, hands holding objects, bent knees, identical objects); as well as distinct indoor and outdoor scenes (e.g., nature, streets, oceans, rooms, toilets; see Fig.~\ref{fig:teaser}). All findings are quantitatively evaluated on measured fMRI not used during interpretation.

\vspace{-0.03cm}
\noindent
\textbf{Our contributions are as follows:}
\vspace{-0.1cm}
\begin{itemize}[leftmargin=*] 
\setlength{\itemsep}{1.5pt}
\item We propose \emph{BrainExplore}, a large-scale, automated framework, that discovers thousands of interpretable fMRI patterns including previously unreported ones.
\item We enrich fMRI decomposition with predicted fMRI, which substantially improves interpretability.
\item We propose an SAE fMRI decomposition that finds interpretable patterns beyond standard methods.
\item We will release the brain-inspired concept dictionary, the large-scale dataset of image–fMRI-explanations ranking, and the code, providing a benchmark for future work. 
\end{itemize}

\vspace{-0.19cm}
\section{Related Works}
\label{sec:related}
\vspace{-0.06cm}

\vspace{-0.12cm}
\paragraph{Contrasting predefined stimulus categories.}
A large body of work has studied category-selective activations by contrasting responses to hand-picked stimulus classes.
For example, researchers have mapped the extrastriate body area (EBA) and related regions using body-selective contrasts~\cite{weiner2011not,downing2001cortical}, and characterized scene-selective regions such as the parahippocampal place area (PPA)~\cite{epstein2019scene,park2011disentangling}.
Other studies have identified face-selective regions including the fusiform face area (FFA)~\cite{kanwisher2006fusiform}, examined overlapping but separable responses to faces and food in fusiform cortex~\cite{Adamson18,Jain23}, or investigated object-related properties~\cite{konkle2010examining}.
These approaches have been highly informative, but they are limited by the need to predefine a small set of categories, to use relatively “clean” images dominated by a single concept, and to test each hypothesis separately.
Moreover, the same voxels and regions can participate in many concepts, making it difficult to discover nuanced or overlapping representations using only category contrasts.

\vspace{-0.32cm}
\paragraph{Decomposing fMRI signals.}
Unsupervised decomposition methods provide an alternative route to fMRI interpretability.
Commonly used techniques include Principal Component Analysis (PCA)~\cite{hotelling1933analysis}, Independent Component Analysis (ICA)~\cite{comon1994independent}, and Non-negative Matrix Factorization (NMF)~\cite{lee1999learning}.
These have been widely applied to fMRI research such as resting-state analysis and connectivity
\cite{Beckmann04,Calhoun01,Smith09,Viviani05,zhong2009detecting,mckeown1998independent}.
Stimulus-driven decompositions for interpretability have also been explored, in domains such as speech and audio~\cite{zengdisentangling,norman2015distinct,deniz2019representation} and visual stimuli~\cite{ecker2007detecting,Huth12,brouwer2009decoding}.
For visual concept discovery, early work used PCA to capture large-scale semantic maps~\cite{ecker2007detecting,Huth12,brouwer2009decoding}, while later studies leveraged NMF to obtain more interpretable components for categories such as bodies and food~\cite{van2024core,zhao2024separate,Khosla22,zhao2024separate}.
ICA has seen more limited use, mainly for demonstrating retinotopic organization and low-level visual mapping~\cite{van2009intrinsic}.
Most of these studies focus on specific regions (e.g., early visual cortex, EBA, PPA) or on a small set of concepts (e.g., food, social vs.\ non-social images), and typically examine a single decomposition method and only the first few components via manual inspection.
We systematically compare multiple decomposition methods for brain interpretability, and scale analysis to tens of thousands of components across many regions and methods.
We also leverage image-to-fMRI models to greatly expand the effective training set and image pool used for decomposition, and we introduce SAE~\cite{cunningham2023sparseautoencodershighlyinterpretable} for fMRI decomposition.

\vspace{-0.35cm}
\paragraph{Image–fMRI models.}
Recent years have brought substantial progress in models that link images and brain activity.
This includes encoding models that predict fMRI responses from images~\cite{Kay08,Naselaris11,beliy2024wisdom}, transformations between subjects or brains~\cite{wasserman2024functional,yamada2015inter}, and image reconstruction from fMRI~\cite{beliy2026brainit,scotti2024mindeye2}.
Beyond pure prediction, several works have used such models to explore cortical representations.
Early studies related voxel activations to predefined semantic concepts, for example in face- and place-selective regions~\cite{ccukur2013functional,ccukur2016functional}.
More recent work has learned functional voxel clusters and visualized the kinds of images each cluster corresponds to~\cite{beliy2024wisdom}.
Another line of work uses image-to-fMRI encoders together with generative image models to synthesize images that strongly activate particular regions or voxels~\cite{hwang2025silico,braindiffusion,neurogen,luo2025brain}.
However, these approaches still operate at the level of individual voxels or entire regions:
regions are too coarse to capture fine-grained mixtures of concepts, while voxels are too local and can participate in many concepts, making nuanced representations difficult to disentangle.
Our approach is complementary.
We use image-to-fMRI models primarily as a tool for data augmentation and interpretability at the \emph{pattern} level:
predicted responses for large image sets greatly expand the pool of images used to learn decompositions and to retrieve maximally activating examples for each component, enabling more robust and fine-grained hypothesis testing.

\vspace{-0.33cm}
\paragraph{Automated interpretability.}
Interpretability of neural networks advanced rapidly in recent years
\cite{bau2017network,hernandez2021natural,gandelsman2023interpreting,gandelsman2024interpreting}.
Recognizing that manual inspection does not scale to modern models, several works have proposed automated interpretability pipelines
\cite{ghorbani2019towards,oikarinen2022clip,schwettmann2023find,shaham2024multimodal}.
The human brain shares many of the same challenges: it is a large, complex system with many units (voxels) and a vast repertoire of concepts.
However, as far as we know, no large-scale, automatic interpretability pipeline has been proposed for the brain. Our work bridges these directions.
We combine unsupervised fMRI decomposition with an automatic interpretability pipeline.
Together with data augmentation from image-to-fMRI models and the use of SAEs, this yields a scalable framework that discovers many interpretable brain patterns and provides a benchmark for evaluating and improving future decomposition methods.

\begin{figure*}[t]
  \centering
  \includegraphics[width=\textwidth]{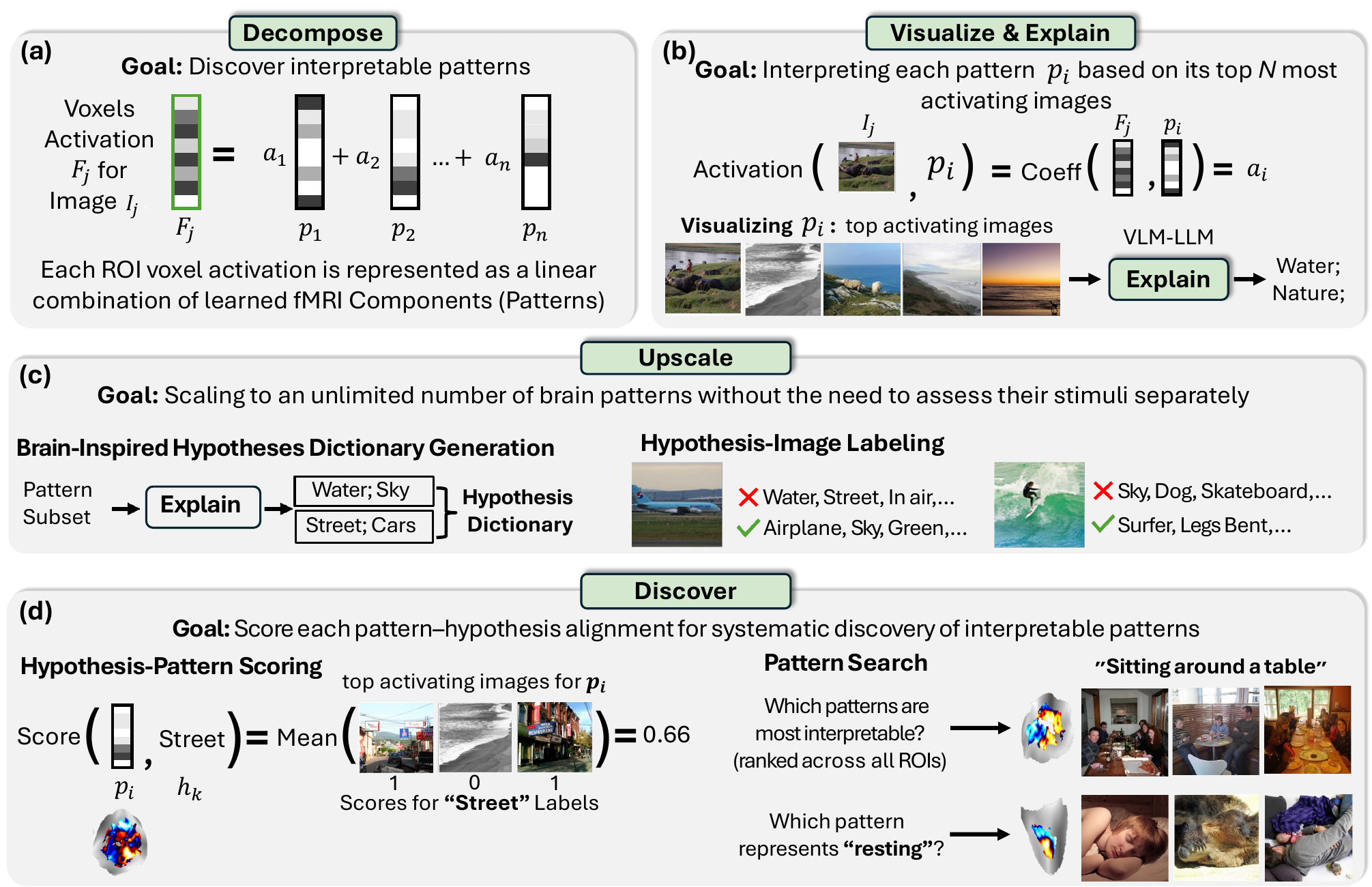}
  \vspace{-0.39cm}
  \caption{\textbf{BrainExplore Framework.}
  \small \textbf{(a) \emph{Decompose:}} per-region fMRI decomposition to discover interpretable patterns (Sec.~\ref{sec:Method_Decomposition}).
  \textbf{(b) \emph{Visualize \& Explain:}} retrieve the top activating images for each pattern and interpret its semantics (Sec.~\ref{sec:Method_Visualize}).
  \textbf{(c) \emph{Upscale:}} scale to an unlimited number of patterns by building a brain-inspired dictionary and labeling each image with respect to each hypothesis (Sec.~\ref{sec:Upscale}).
  \textbf{(d) \emph{Discover:}} score each pattern–hypothesis alignment, enabling systematic discovery of the most interpretable patterns and the pattern that best explains any given hypothesis (Sec.~\ref{sec:Method_Discover}).}
  \vspace{-0.47cm}
  \label{fig:methods}
\end{figure*}

\vspace{-0.3cm}
\section{Methods}
\label{sec:Methods}
\vspace{-0.29cm}

We first describe the experimental setting used in our framework (Sec.~\ref{sec:exp_details}). We then present \emph{\textbf{BrainExplore}}, our brain interpretability framework with four parts (see Fig.~\ref{fig:methods}): 
(i) \emph{\textbf{Decompose}}: per-region fMRI decomposition into pattern components (Sec.~\ref{sec:Method_Decomposition});
(ii) \emph{\textbf{Visualize \& Explain}}: retrieve the top activating images for each pattern and provide an interpretation of its semantics (Sec.~\ref{sec:Method_Visualize});
(iii) \emph{\textbf{Upscale}}: scale to an unlimited number of brain patterns without interpreting each stimulus separately (Sec.~\ref{sec:Upscale});
(iv) \emph{\textbf{Discover}}: score each pattern–hypothesis alignment for systematic discovery of interpretable patterns (\cref{sec:Method_Discover}).

\vspace{-0.28cm}
\subsection{Experimental Setting}
\label{sec:exp_details}
\vspace{-0.20cm}

We use the Natural Scenes Dataset (NSD)~\citep{allen2022massive}, a 
large publicly available 7-Tesla fMRI dataset that records responses from 8 subjects viewing diverse natural images drawn from COCO~\citep{lin2014microsoft}. 
The dataset contains $\sim$73k image–fMRI pairs in total, with around 10k images per subject (some images are shared across subjects).
We adopt the post-processed version of NSD released by \citet{gifford2023algonauts}, which includes $\sim$40k voxels per subject together with a predefined division into mainly vision-related ROIs. 
These ROIs are used for our per-ROI decomposition and subsequent analyses. To enrich the training data, we additionally use predicted fMRI responses for images not viewed by any subject.
Specifically, we take $\sim$120K additional natural images from the unlabeled portion of COCO and employ the image-to-fMRI encoder of \citet{beliy2024wisdom} to predict fMRI responses for each subject.
This augmentation produces a substantially larger set of image–fMRI pairs, which we use both for training decompositions and for retrieving maximally activating images.  
Importantly, all interpretations are evaluated and verified on measured fMRI.

\begin{figure*}[t]
  \centering
\includegraphics[width=0.98\textwidth]{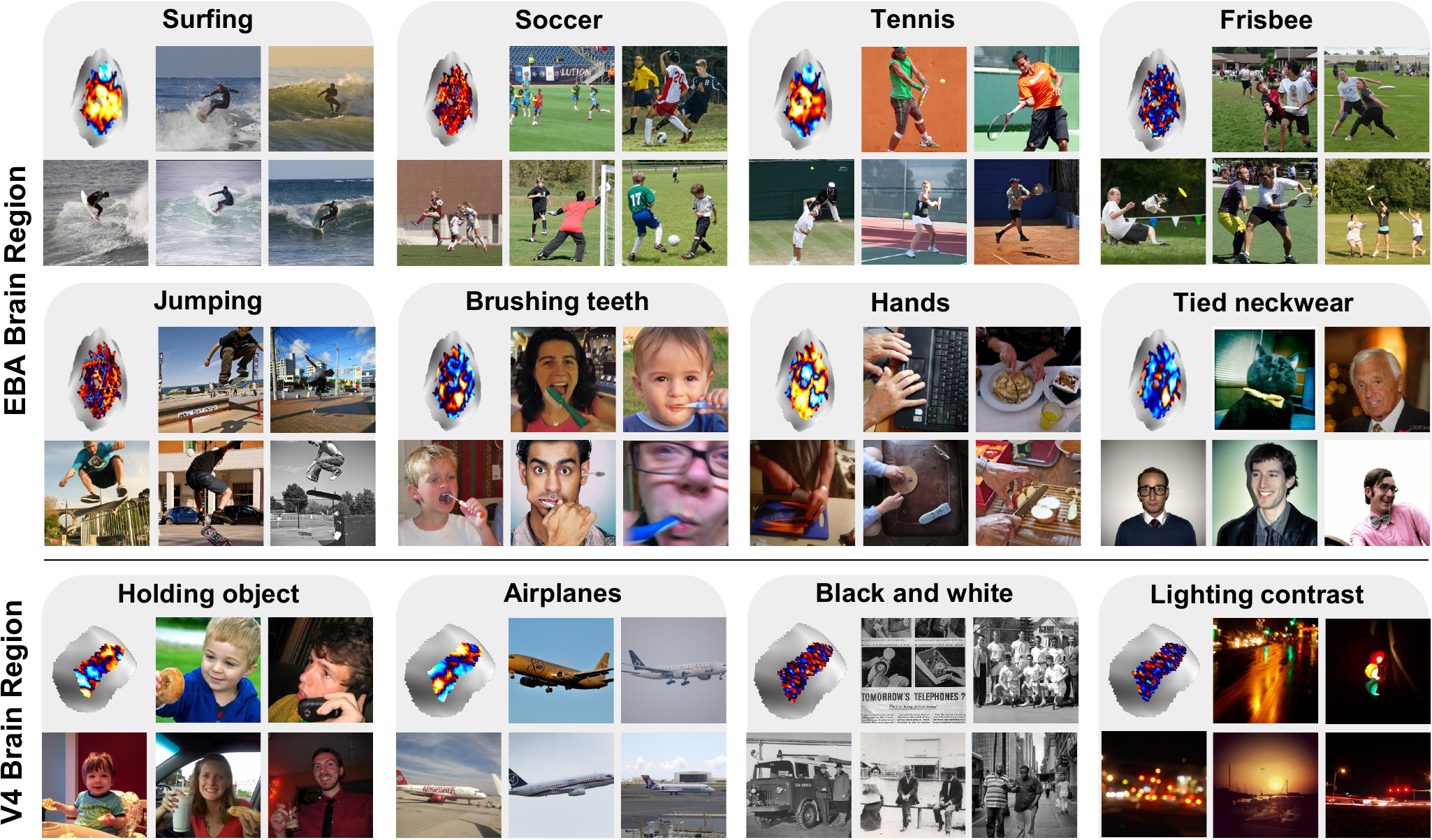}
  \vspace{-0.1cm}
    \caption{\textbf{Discovered Interpretable Patterns (EBA and V4).} 
    \small We show patterns of subject 1 with their top activating images and selected explanations. 
    EBA is known to process bodies and actions; V4 is known to encode mid-level features (e.g., color, shape).}
  \vspace{-0.58cm}
  \label{fig:results}
\end{figure*}

\vspace{-0.15cm}
\subsection{fMRI Decomposing}
\label{sec:Method_Decomposition}
\vspace{-0.26cm}

The first step in our pipeline performs per-ROI decomposition of fMRI activity to extract component patterns that may correspond to meaningful visual representations.
Conceptually, each decomposition method learns a set of fMRI patterns (components) such that any fMRI response in a given ROI can be approximated as a linear combination of them (\cref{fig:methods}a).
Each trained decomposition yields two outputs:
\emph{(i) a component matrix}, where each column is a learned fMRI pattern (for PCA, NMF, and ICA this is the standard loading matrix; for SAEs it corresponds to the decoder weights); and
\emph{(ii) a coefficient matrix}, which contains the linear coefficients reconstructing each fMRI response from these patterns.
Importantly, all decompositions are learned purely from fMRI responses; no image features or labels are used, ensuring that all inferred visual representations arise from brain activity rather than externally imposing image semantics.

We train three standard decomposition methods (PCA, NMF, and ICA) as well as SAEs.
For each method, we train two variants:
(i) using only measured fMRI responses ($\sim$10k responses per subject), and
(ii) using the combined dataset of measured fMRI and 120k predicted responses.
For each method, we adopt default parameter settings and vary a few hyperparameters 
(e.g., variance thresholds for PCA/NMF/ICA and different seeds; sparsity settings and expansion factors for SAE). 
Each decomposition produces a large pool of candidate patterns per ROI, later used in the \emph{Discover} step.
As a baseline, we also treat individual voxel activations as a single-component “decomposition”, effectively serving as a one-hot voxel basis. Training details are provided in (\cref{sec_sup:SAE_training}).

\vspace{-0.28cm}
\subsection{From fMRI Patterns to Visual Concepts}
\label{sec:Method_Visualize}
\vspace{-0.2cm}

To interpret each decomposed brain pattern, we first connect it to the set of stimuli that drive it (\cref{fig:methods}b). Each fMRI response is directly associated with a stimulus --- the natural image viewed by the subject.
After decomposition, every fMRI response is represented by a vector of coefficients, one per pattern component, indicating how strongly each pattern contributes to its reconstruction.
We assume that the responses with the highest coefficients for a given pattern are the most informative about its underlying visual semantics.
Thus, for every pattern, we select its top-\(N\) activating responses and collect their corresponding images, producing a set of images that \emph{visualize} what most strongly activates that pattern. We perform this separately for measured-fMRI responses and predicted responses.
To \emph{explain} each pattern, we take the six most activating images from the measured-fMRI pool and the ten most activating images from the predicted-fMRI pool (16 images total).
We generate captions for these images using a Vision–Language Model (VLM) and then prompt an LLM to synthesize 5–10 hypotheses capturing what is shared across them 
(e.g., object identity, pose, texture, scene category). These hypotheses serve as candidate semantic explanations for the pattern.

\begin{figure*}[t]
  \centering
\includegraphics[width=0.98\textwidth]{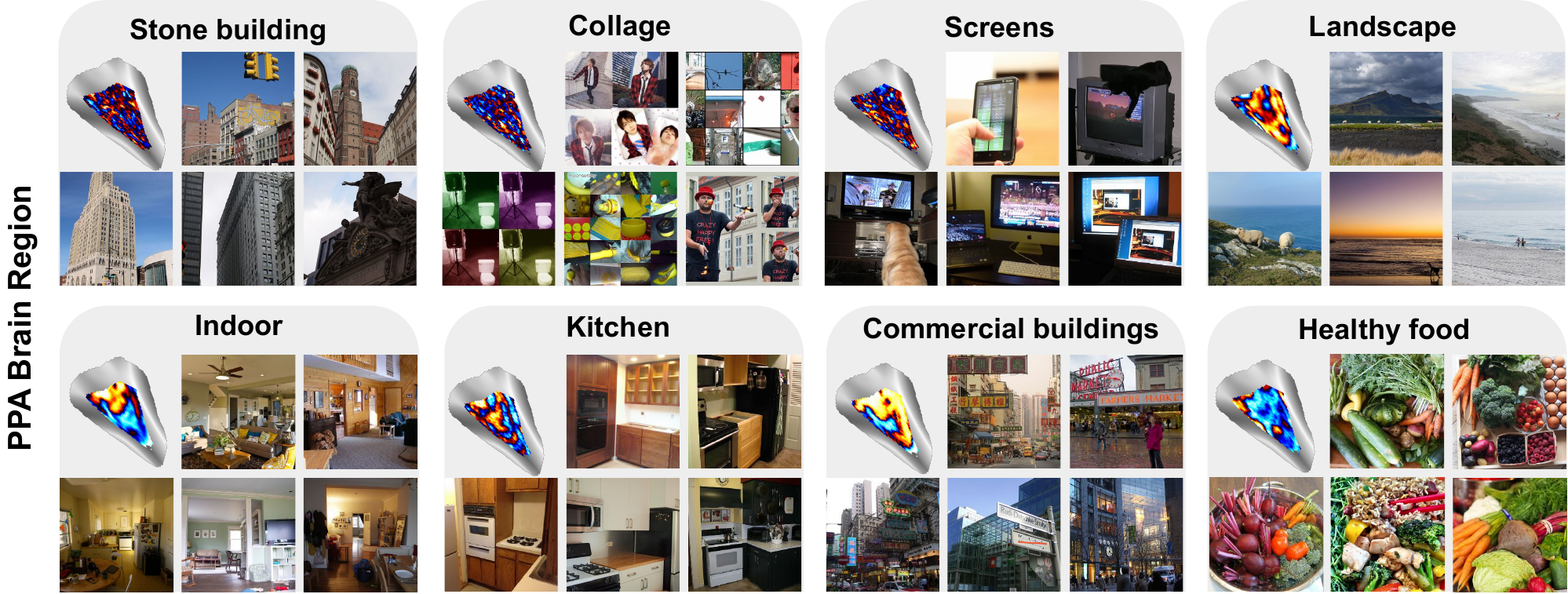}
  \vspace{-0.08cm}
 \caption{\textbf{Discovered Interpretable Patterns (PPA).}
\small We show Subject~1 patterns with top activating images and selected explanations. PPA encodes scenes and places (e.g., indoor/outdoor, landmarks).}
  \vspace{-0.43cm}
  \label{fig:results2}
\end{figure*}

\vspace{-0.08cm}
\subsection{Scaling to an Unlimited Number of Patterns}
\label{sec:Upscale}
\vspace{-0.17cm}

Explaining every component in every ROI and decomposition method is costly and inefficient, especially with high-dimensional SAEs, multiple hyperparameters and methods. To scale interpretability, we introduce two steps~(\cref{fig:methods}c): (i) \textbf{\emph{Hypothesis dictionary generation}}, which constructs a brain-activity-inspired set of candidate hypotheses; and (ii) \textbf{\emph{Hypothesis–image labeling}}, which assigns binary labels for each hypothesis and image. Together, these two steps create large-scale associations between visual stimuli and candidate descriptions of brain activity, independent of any specific pattern.
Once the dictionary and image labels are in place, any new brain pattern can be evaluated by measuring how consistently its top-activating images express concepts from the dictionary.

\vspace{-0.35cm}
\paragraph{Hypothesis dictionary generation.}
We first identify patterns most likely to be interpretable.
To do so, we compute a CLIP-based \emph{consistency} score for every pattern: we embed its top activating images, compute pairwise similarities, and use their mean as a proxy to semantic coherence. 
We then select the top 40 patterns per ROI and decomposition method and run the \emph{Visualize \& Explain} step on this subset ($>$ 10k patterns), yielding more than 1k unique candidate explanations. These are aggregated into a brain-inspired dictionary of visual concepts.
To remove duplicates 
we embed each hypothesis with a BGE text encoder~\cite{chen2024bge} and merge concepts with high embedding similarity. This produces a dictionary of \(\sim\)1{,}300 concepts that we can use to label all stimuli offline with respect to every candidate concept, creating large-scale associations between images and candidate brain activity descriptions, independent of any specific pattern. Further details and prompts appear in \cref{sec_sup:additional_details}.

\vspace{-0.35cm}
\paragraph{Hypothesis–image labeling.}
Our goal is to obtain, for every image in the dataset (both for images with measured and predicted fMRI), a binary vector indicating which concepts from the dictionary apply to that image.
Directly querying a VLM with all 1{,}300 hypotheses per image is infeasible and often leads to inaccurate results, 
therefore we use a two-stage procedure.
First, we use CLIP to shortlist a small set of candidate hypotheses per image:
We embed each hypothesis under multiple prompt templates, compute its similarity to the CLIP image embedding, and retain the top 300 hypotheses per image.
Second, we perform VLM-based verification.
For each shortlisted candidate hypothesis, we prompt a VLM to output a binary decision (0/1) indicating whether the hypothesis holds for a given image.
To improve reliability, we run a second verification pass with a different prompt for all hypotheses initially marked as positive, and keep only those confirmed twice.
This produces a sparse but high-quality binary image–hypothesis label matrix, used later for large-scale pattern–hypothesis scoring. Additional details and labeling examples are provided in \cref{sec_sup:additional_details}.

\begin{table}[ht] \centering \setlength{\tabcolsep}{6pt} \small 
\caption{\textbf{Average Number of Interpretable Hypotheses.} \small The number/fraction of hypotheses with an alignment score above a threshold (0.5 or 0.8) for at least one pattern, averaged across 8 subjects.}
\vspace{+0.12cm}
\begin{tabular}{@{}l*{4}{c}@{}} \toprule & \multicolumn{2}{c}{Measured fMRI} & \multicolumn{2}{c}{+ Predicted fMRI} \\ \cmidrule(lr){2-3}\cmidrule(lr){4-5} Method & $>0.5$ & $>0.8$ & $>0.5$ & $>0.8$ \\ \midrule Voxels & $42$ / $3.17\%$ & $4$ / $0.34\%$ & $86$ / $6.50\%$ & $15$ / $1.19\%$ \\ PCA & $20$ / $1.53\%$ & $4$ / $0.32\%$ & $78$ / $5.90\%$ & $11$ / $0.84\%$ \\ NMF & $19$ / $1.45\%$ & $3$ / $0.23\%$ & $61$ / $4.62\%$ & $10$ / $0.78\%$ \\ \addlinespace[2pt] ICA & $11$ / $0.88\%$ & $2$ / $0.22\%$ & $207$ / $15.56\%$ & $34$ / $2.62\%$ \\ \addlinespace[2pt] SAE & $\mathbf{104}$ / $\mathbf{7.87\%}$ & $18$ / $1.33\%$ & $219$ / $16.46\%$ & $51$ / $3.80\%$ \\ \addlinespace[2pt] SAE+ICA & $105$ / $7.86\%$ & $\mathbf{18}$ / $\mathbf{1.34\%}$ & $\mathbf{250}$ / $\mathbf{18.81\%}$ & $\mathbf{58}$ / $\mathbf{4.36\%}$ \\ \bottomrule \label{tab:hypothesis} \end{tabular} 
\vspace{-0.75cm} \end{table}

\vspace{-0.23cm}
\subsection{Discover Interpretable Brain Patterns}
\label{sec:Method_Discover}
\vspace{-0.2cm}

The goal of the \emph{Discover} step (\cref{fig:methods}d) is to determine which visual concept each fMRI pattern represents.
It uses the outputs of \emph{Upscale} (the hypothesis dictionary and image–hypothesis labels) to evaluate any decomposition component pattern by checking whether the images that activate it express consistent visual concepts.

\vspace{-0.35cm}
\paragraph{Hypothesis–pattern scoring.}
For each pattern, we quantify how strongly it aligns with each candidate hypothesis.
Every fMRI response has a coefficient indicating how much a given pattern contributes to reconstructing that response. We rank all images by this coefficient and select the top \(0.2\%\) most activating images for that pattern. This is done separately for the measured and predicted fMRI pools, excluding responses that do not meaningfully activate the pattern (SAE coefficients \(<0.01\); non-SAE coefficients \(\leq 0\)). Since each image has binary labels for all hypotheses, we can directly measure how well a hypothesis matches the pattern.
For a given pattern and hypothesis \(h\), we compute its score as \(\frac{\#\{\text{top-activating images labeled with } h\}}{N},
\) where \(N\) is the total number of selected top-activating images.
This measures how consistently concept \(h\) appears among the images that drive the pattern. Some hypotheses are globally rare, so even a small number of positives may indicate strong alignment. To avoid penalizing such cases, we apply a normalization factor based on the global frequency of the hypothesis (the proportion of images labeled ``1'' for that hypothesis across the dataset), capping the factor at~2.
The resulting normalized score reflects how unexpectedly frequent the hypothesis is within the pattern’s top images. Scores are computed independently for the measured and predicted fMRI pools, and the final hypothesis-pattern score is defined as their average.

\vspace{-0.35cm}
\paragraph{Pattern search.}
Given the pattern–hypothesis scores, we support two complementary searches:
(i) \emph{pattern search}: find the most interpretable pattern(s); and
(ii) \emph{hypothesis search}: for a given hypothesis (e.g., “open mouth”), find the best-explained pattern.
Both searches can be run \emph{per ROI} (best pattern within an ROI) or \emph{across ROIs} (best pattern over all ROIs), enabling region-specific (\emph{if/where} a hypothesis is represented) and method-level (\emph{which} decompositions capture it) analyses.

\vspace{-0.25cm}
\section{Experiments}
\label{sec:results}
\vspace{-0.22cm}
\subsection{Evaluation Protocol}
\vspace{-0.23cm}

To ensure fair evaluation, we split each pool into disjoint \emph{ranking} and \emph{evaluation} halves: the ranking set is used to assign best pattern-hypothesis pairs; the evaluation set is used to report final metrics at Sec. \ref{sec:results}. 
In the main paper, we present results for Subject~1 from the NSD dataset. Quantitative and qualitative results for the other subjects are provided in \cref{sec_sup:additional_Quantitative,sec_sup:additional_Visualizations}, demonstrating the effectiveness of our pipeline across subjects.
Since every pattern receives a score, searches can be conducted not only within a single decomposition, but also \emph{across} methods and hyperparameters, allowing integrated comparison and selection. 


\vspace*{-0.25cm}
\subsection{Quantitative Evaluation}
\vspace*{-0.22cm}

We report two metrics:
(i) \textbf{\emph{Number of Interpretable Hypotheses}} (Table~\ref{tab:hypothesis}): the number/fraction of hypotheses that achieve an alignment score above a specified threshold (0.5 or 0.8) with at least one pattern. This can be computed either per ROI or across all ROIs.
(ii) \textbf{\emph{Number of Interpretable Patterns}} (Appendix \cref{tab:patterns}): the number of patterns whose best hypothesis alignment exceeds the same threshold (0.5 or 0.8). To avoid counting near-duplicate patterns, we remove any component whose voxel-wise correlation with a higher-scoring pattern exceeds 0.5. This table shows that SAE yields far more interpretable patterns than other methods: 
approximately 9k patterns across the brain at a 0.5 threshold, 
compared to \(\sim\)6k for the Voxels baseline and 226 for ICA. 

\vspace*{-0.27cm}
\paragraph{Predicted fMRI substantially improves interpretability.}
Table~\ref{tab:hypothesis} compares decomposition methods trained on measured fMRI alone versus augmented with predicted fMRI, used both to learn the decomposition and to retrieve top-activating images. Predicted fMRI yields large gains across methods: ICA increases from an average of around 12 to 207 interpretable hypotheses at the 0.5 threshold, and SAE increases from around 105 to 218. 

\vspace{-0.27cm}
\paragraph{SAE provides the largest number of interpretable hypotheses.}

{SAE yields the highest number of interpretable hypotheses, both with and without predicted fMRI.}

\vspace{-0.27cm}
\paragraph{Combining decompositions yields the best performance.}
 Integrating across methods (ICA+SAE) yields the strongest performance. This is enabled by our automatic scoring, which evaluates any number of pattern–hypothesis pairs and supports aggregation across decompositions.

\begin{figure}[t]
  \centering
  \begin{minipage}[t]{0.51\textwidth}
    \centering
    \includegraphics[width=\textwidth]{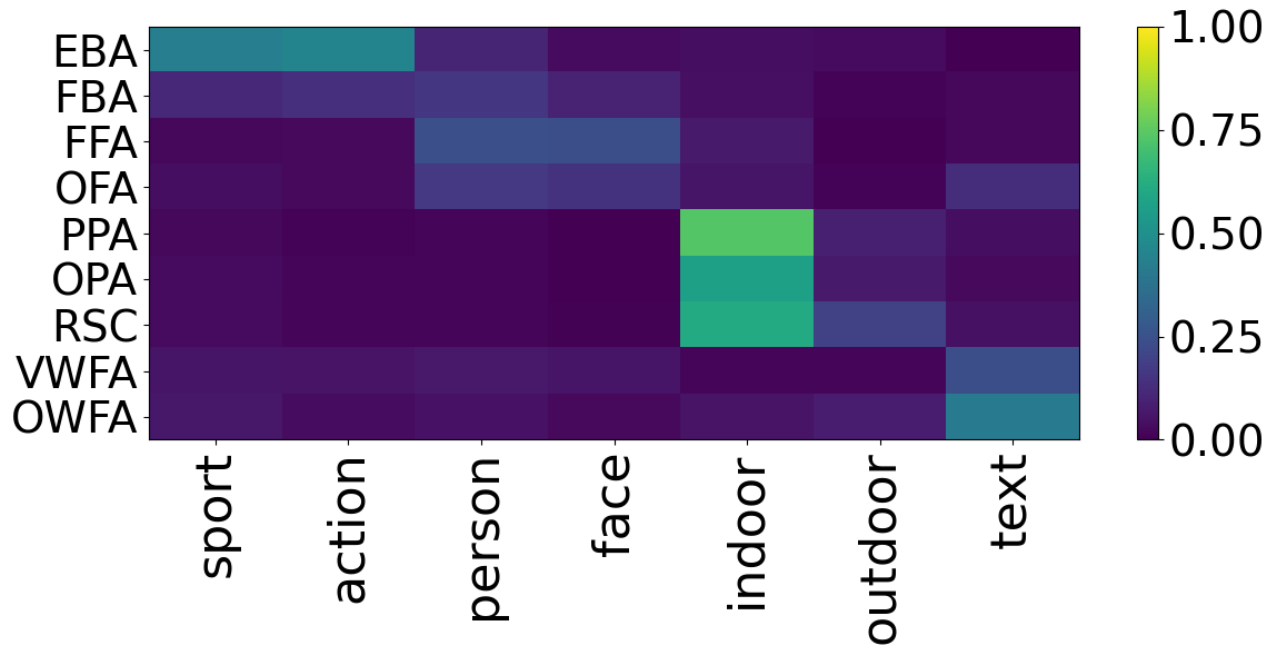}
    \vspace{-0.4cm}
\captionof{figure}{\textbf{ROI--concept alignment.} \small Applying our pipeline to average ROI activation shows scores consistent with the known functional selectivity of visual regions. For example, FFA/OFA align with faces and place regions (e.g. PPA) with indoor/outdoor scenes.}
    \label{fig:ROI–concept alignment}
  \end{minipage}%
  \hfill
  \begin{minipage}[t]{0.46\textwidth}
    \centering
    \setlength{\tabcolsep}{8pt}
    \small
    \vspace{-3.84cm}
    \begin{tabular}{@{}lccc@{}}
    \toprule
    ROI & $k=2$ & $k=4$ & $k=8$ \\
    \midrule
    EBA  & 0.84* & 0.72* & 0.61* \\
    FBA  & 0.76  & 0.60* & 0.48* \\
    FFA  & 0.73  & 0.55  & 0.42* \\
    OFA  & 0.69  & 0.50  & 0.38  \\
    PPA  & 0.75  & 0.58* & 0.44* \\
    OPA  & 0.80* & 0.66* & 0.55* \\
    RSC  & 0.67  & 0.47  & 0.35  \\
    VWFA & 0.73  & 0.55  & 0.41  \\
    OWFA & 0.70  & 0.51  & 0.38*  \\

    \bottomrule
    \end{tabular}
    \vspace{-0.05cm}
\captionof{table}{\textbf{Concept consistency across subjects.} \small We collected the interpretable concepts discovered in each ROI and measured their overlap across subject subsets of size $k$, compared to a null baseline. Asterisks indicate $p<0.05$.}
    \label{tab:Concept consistency across subjects.}
  \end{minipage}
  \vspace{-0.5cm}
\end{figure}

\vspace{-0.25cm}
\subsection{Alignment with Established Neuroscientific Knowledge}
\vspace{-0.2cm}

Our method reveals a substantially larger number of interpretable patterns and visual concepts than prior work. These include previous reported findings, concepts aligned with known functionality of specific brain regions, but to the best of our knowledge not previously shown, and additional less expected concepts, reflecting that many aspects of cortical organization remain not fully understood.
We next validate our results by showing  consistency with established neuroscientific knowledge.

\vspace{-0.2cm}
\paragraph{Alignment with known category selectivity in functional regions.}
We conducted two experiments to validate alignment between our discovered concepts and established functional selectivity of visual regions. 
(i) \textbf{ROI--concept alignment:} We applied our scoring procedure at the ROI level, using the average activation of each ROI instead of individual voxels or decomposition components, to compute an ROI--concept alignment score. Since ROIs are known to be polysemantic, we do not expect very high alignment scores. Nevertheless, we expect the dominant concepts in each ROI to reflect its known functional role. As shown in Fig.~\ref{fig:ROI–concept alignment}, the resulting alignments are consistent with established 
knowledge: for example, EBA aligns strongly with actions and sports~\cite{weiner2011not,downing2001cortical}; FFA aligns with people and faces~\cite{kanwisher2006fusiform}; PPA, OPA, and RSC align with locations~\cite{epstein2019scene,ccukur2016functional}; and OWFA and VWFA align with text-related concepts~\cite{cohen2000visual,mccandliss2003visual}. 
(ii) \textbf{LLM-based relevance labeling:} We further automatically labeled each concept--ROI pair according to whether it is generally consistent with the known functional role of that ROI. We then measured, for each ROI, the fraction of concepts labeled as relevant that were found to be interpretable in that ROI. The results, shown in Fig.~\ref{fig:concept_roi_alignment} and discussed in more detail in Appendix ~\ref{sec_sup:extended_validation}, further support the validity of the discovered patterns.

\vspace{-0.3cm}
\paragraph{Consistency across subjects.}
The functional roles of visual regions are expected to show some consistency across subjects. To evaluate this, for each ROI and subject, we collected the set of interpretable concepts discovered in that ROI and measured their overlap across subjects. In Table~\ref{tab:Concept consistency across subjects.}, we report the fraction of concepts in each ROI consistent across all subject pairs, and groups of 4 and 8 subjects. As expected, consistency is high but not perfect, reflecting inter-subject variability. We further analyze the cross-subject consistency of patterns in Appendix ~\ref{sec_sup:cross_subject_pattern} and Fig.~\ref{fig:IoU_across_subjects}.

\vspace{-0.15cm}
\subsection{Visual Representations in
the Human Brain}
\vspace{-0.15cm}
\paragraph{BrainExplore reveals fine-grained patterns across the brain.}

BrainExplore identifies interpretable patterns effectively.
For any hypothesis of interest, we can retrieve the best aligned pattern within a specific ROI or across ROIs, and visualize its representation via the top activating images.
\cref{fig:teaser,fig:results} show examples from the top 16 images per pattern (8 from measured, 8 from predicted fMRI; full grids appear in \cref{sec_sup:additional_Visualizations}).
Although the predicted fMRI pool is much larger and often yields clearer visualizations, we validate each explanation on measured fMRI as well.
Overall, BrainExplore uncovers many fine-grained patterns, including ones not previously identified. 
For example, in EBA, classically linked to body perception and action, we find patterns selective for specific sports (surfing, soccer, tennis, frisbee), actions (jumping, tooth brushing), and body oriented concepts (hands, neckwear).
In PPA, known to respond to scenes, we observe a more nuanced division than prior indoor-outdoor contrasts, including distinct outdoor concepts for landscapes, commercial buildings, and stone architecture. Additional concepts and ROIs are shown in \cref{sec_sup:additional_Visualizations}.

\begin{figure*}[t]
  \centering
\includegraphics[width=0.96\textwidth]{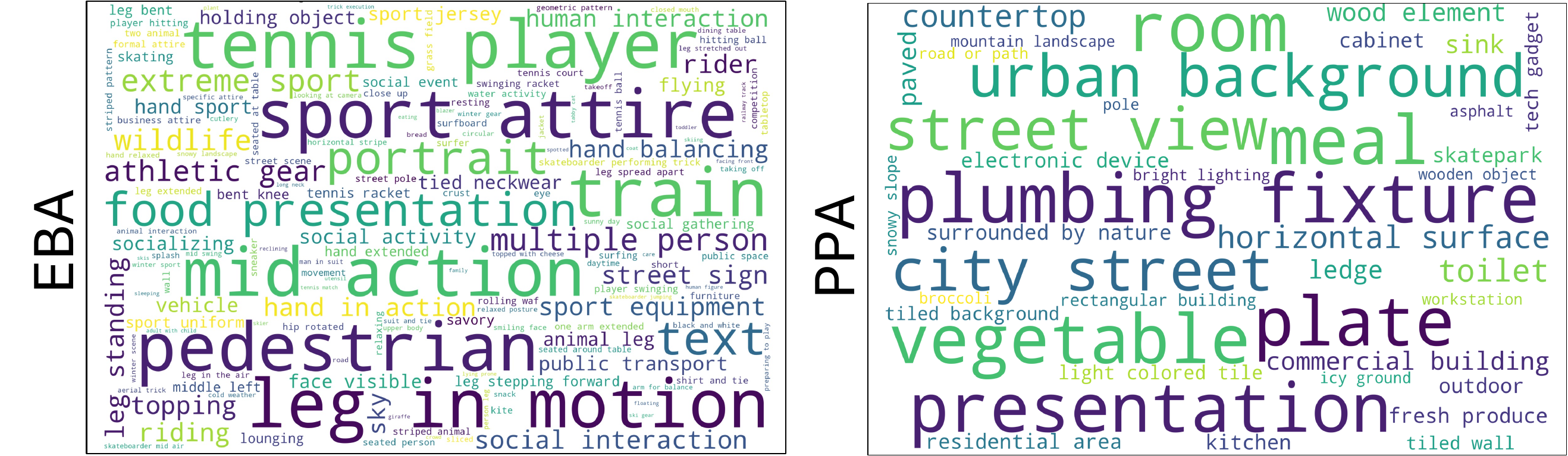}
  \vspace{-0.1cm}
\caption{\textbf{Concepts best explained by each ROI (EBA, PPA).}
\small Each concept is assigned to a single ROI—the one with the highest alignment score. 
Only concepts with alignment \(>0.5\) are shown; word size reflects the alignment score within the assigned ROI.}
  \vspace{-0.52cm}
  \label{fig:words}
\end{figure*}

\vspace*{-0.25cm}
\paragraph{Different ROIs represent different semantics.}
Different ROIs are known to support distinct functions. Many of our discovered patterns align strongly with these known roles, which both validates our approach and lends credibility to novel findings. Localizing established functions remains valuable—especially when we reveal finer semantics. To examine this systematically, we first find, for each hypothesis, the best matching pattern across ROIs; then, for each ROI, we list the hypotheses it best explains. In \cref{fig:words}, we show examples where word size reflects the hypothesis–pattern score in that ROI. As shown, our method recovers many concepts, with different ROIs best explaining different ones, as expected.

\vspace{-0.15cm}
\paragraph{Interpretable patterns are more localized with SAE.}
The brain exhibits meaningful spatial organization at both regional and subregional scales. While low-level maps are well studied, higher-level semantic organization remains less clear. We find that SAEs yield interpretable patterns that are notably more spatially localized, despite receiving only 1D voxel vectors and no spatial information or spatial constraint. In \cref{fig:results}, we show patterns from both ICA and SAE. In EBA, for example, \emph{soccer}, \emph{frisbee}, and \emph{jumping} are derived using ICA, whereas the remaining patterns come from SAE. Additional side-by-side comparisons of pattern localization appear in \cref{sec_sup:additional_Visualizations}. SAE patterns are visibly more compact and clustered rather than scattered. Greater localization is both biologically plausible and of strong interest to the neuroscience community.

\vspace{-0.25cm}
\subsection{Ablations \& Analysis}
\vspace{-0.15cm}

In \cref{sec_sup:Ablation} we analyze several aspects of our pipeline, including:
(i) \textbf{\emph{SAE ablations}} - the importance of sparsity and of expanding dimensionality;
(ii) \textbf{\emph{Predicted fMRI pool size}} - results with varying image–fMRI pool sizes for retrieving top-activating images and their effect on interpretability;
(iii) \textbf{\emph{Per-ROI quantitative results}} - detailed metrics reported per ROI; and
(iv) \textbf{\emph{Complementarity across decompositions}} - the gains from combining different methods and an analysis of concepts uniquely captured by each.

\vspace{-0.25cm}
\section{Discussion \& Limitations}
\label{sec:Conclusion}
\vspace{-0.25cm}

We introduced \emph{BrainExplore}, a scalable framework for discovering visual concept representations across the visual cortex. By combining multiple decomposition methods within an automatic pipeline and enriching both training and retrieval with a large pool of predicted fMRI signals, BrainExplore reveals thousands of interpretable patterns capturing fine-grained visual concepts across regions. Despite the advantages of our large-scale pipeline, several limitations remain. Our approach uses VLM-based labeling, which can be noisy and may reduce the accuracy of the resulting scores. We mitigate this through a careful two-stage procedure, performing one-concept-at-a-time labeling followed by a second VLM verification step. Moreover, while our extensive evaluation supports the validity of the discovered patterns, definitive validation of newly identified representations will require future targeted neuroscience experiments. We hope that \emph{BrainExplore} will help guide the design of such experiments.
Overall, our work takes a meaningful step toward understanding visual representations in the brain and provides a practical tool for large-scale discovery. It uncovers multiple previously unreported visual representations and can help guide future computational and experimental studies.

\section*{Acknowledgments}
This research was partially supported by the European Research Council (ERC) under the Horizon programme, grant number 101142115. We are also grateful for the support of ARL, MIT-IBM Watson AI Lab, Hyundai Motor Company and ONR MURI.

\bibliographystyle{unsrtnat}
\bibliography{main}

\newpage
\appendix
\begin{center}
    {\LARGE \bfseries Appendix}
\end{center}
\renewcommand{\thefigure}{S\arabic{figure}}  
\renewcommand{\thetable}{T\arabic{table}}    

\setcounter{figure}{0} 
\setcounter{table}{0}

\section{Ablations and Additional Results}
\label{sec_sup:Ablation}

\vspace{-0.1cm}
\paragraph{SAE ablations.}
The two main hyperparameters in our SAEs are (i) the latent-space dimensionality, controlled by the expansion factor, and (ii) the sparsity coefficient, which weights the sparsity regularization on the latent codes. Sparsity encourages each fMRI sample to be reconstructed using only a small number of patterns (i.e., a sparse linear combination of decoder columns).
We evaluate four expansion factors: 0.5, 1, 2, and 4. The settings 0.5 and 1 correspond to “regular” autoencoders that reduce or roughly preserve the dimensionality, while 2 and 4 yield overcomplete representations that expand it. We also evaluate four sparsity coefficients: 0, 1, 2, and 4, where 0 corresponds to no sparsity penalty (a standard autoencoder).

In \cref{tab_sup:sae_ablations} we report, for each configuration, the rank score used for model selection (the evaluation results reported in the main paper are computed on a separate held-out set), together with the percentage of interpretable hypotheses above two thresholds (0.5 and 0.8). Sparsity has a clear and consistent effect: higher sparsity generally produces more interpretable patterns. This is expected, since enforcing that each fMRI sample is reconstructed from only a few patterns encourages those patterns to capture meaningful structure rather than overfitting. 
The expansion factor shows a similar trend: larger expansion factors tend to perform better, with 0.5 performing noticeably worse in most cases. Overall, both hyperparameters are important for SAE interpretability. Expansion factors larger than 4 did not yield consistent additional gains and were computationally expensive. Similarly, increasing the sparsity coefficient beyond 4 did not lead to clear improvements.

\begin{table}[h]
\centering
\small
\setlength{\tabcolsep}{5pt}

\caption{\textbf{SAE Ablation by Expansion Factor and Sparsity.} Percentage of Interpretable Hypotheses. The fraction
of hypotheses that achieve an alignment score above a threshold
 (0.5 or 0.8) with at least one pattern (rank score).}
 \vspace{0.1cm}
\label{tab_sup:sae_ablations}

\begin{tabular}{c c cccc}
\toprule
& & \multicolumn{4}{c}{\textbf{Sparsity}} \\
\cmidrule(lr){3-6}
\textbf{Threshold} & \textbf{Exp. Factor} & 0 & 1 & 2 & 4 \\
\midrule
\multirow{4}{*}{$>0.5$}
  & 0.5 & 12.1\% & 19.2\% & 18.2\% & 22.0\% \\
  & 1   & 14.4\% & 18.6\% & 19.2\% & 22.5\% \\
  & 2   & 16.8\% & 19.8\% & 20.9\% & 22.5\% \\
  & 4   & 18.2\% & 21.1\% & 23.3\% & 25.0\% \\
\midrule
\multirow{4}{*}{$>0.8$}
  & 0.5 & 1.1\% & 3.0\% & 2.6\% & 3.6\% \\
  & 1   & 1.4\% & 2.6\% & 2.9\% & 3.6\% \\
  & 2   & 1.7\% & 3.8\% & 3.5\% & 3.7\% \\
  & 4   & 2.2\% & 2.9\% & 3.5\% & 3.8\% \\
\bottomrule
\end{tabular}

\vspace{-0.35cm}
\end{table}

\vspace{-0.2cm}
\paragraph{Predicted fMRI pool size.}
We aim to disentangle the influence of the retrieval pool size (the pool from which the top-\emph{N} most activating images are retrieved) on interpretability performance. In \cref{tab_sup:pool_size} we report results for retrieval pools enriched with different amounts of predicted fMRI. All decomposition methods are trained on the same combined data (measured + predicted fMRI).
Unlike the main paper, where we also compare models trained only on measured fMRI, here we fix the training data and vary only the retrieval pool. Importantly, increasing the pool size also increases the number of images used to compute the hypothesis scores, since the number of top activating images is defined as a fraction of the pool (0.2\%). Therefore, larger pools not only tend to yield higher interpretability scores, but also make the scores more robust. Even when scores are similar, we therefore prefer larger pools. As shown in the table, enriching the pool with predicted fMRI leads to a clear improvement compared to using measured fMRI alone, and for the best-performing model the largest pool yields the highest and most stable interpretability scores.

\begin{table}[h]
\centering
\setlength{\tabcolsep}{5pt}
\small
\caption{\textbf{Percentage of Interpretable Hypotheses (different retrieval pools).}
All decomposition methods are trained on the same combined data (measured + predicted fMRI). 
Unlike the main paper, where we also compare models trained only on measured fMRI, here we fix the training data and vary only the retrieval pool. 
We report results for pools enriched with different amounts of predicted fMRI (threshold $>$ 0.5).}
\label{sup_tab:hypothesis}
\vspace{0.1cm}
\begin{tabular}{@{}l*{5}{c}@{}}
  \toprule
  Method &
  \shortstack{Measured\\fMRI} &
  \shortstack{+Pred\\30k} &
  \shortstack{+Pred\\60k} &
  \shortstack{+Pred\\90k} &
  \shortstack{+Pred\\120k} \\
  \midrule
  Voxels         & 3.8\% & 7.1\% & 6.8\% & 7.1\% & 6.7\% \\
  PCA (Single)   & 1.8\% & 7.1\% & 7.7\% & 7.7\% & 7.4\% \\
  NMF (Single)   & 2.4\% & 5.6\% & 5.8\% & 5.9\% & 5.5\% \\
  \addlinespace[2pt]
  ICA (Single)   &10.4\% & 18.0\% & 18.5\% & 17.7\% & 18.1\% \\
  ICA (Multiple) & 10.0\% & 17.4\% & 17.6\% & 17.9\% & 18.3\% \\
  \addlinespace[2pt]
  SAE (Single)   & 5.6\% & 16.0\% & 14.6\% & 15.5\% & 15.7\% \\
  SAE (Multiple) & 8.2\% & 17.8\% & 16.8\% & 17.4\% & 17.4\% \\
  \addlinespace[2pt]
  SAE+ICA        & 9.8\% & 20.7\% & 20.6\% & 20.9\% & 21.5\% \\
  \bottomrule
  \label{tab_sup:pool_size}
\end{tabular}
\vspace{-0.4cm}
\end{table}

\paragraph{Per-ROI quantitative results.}
We report, for each ROI, the percentage of explained hypotheses above a 0.5 threshold for the averaged results shown in the paper. V1–V3 are averaged across ventral and dorsal partitions, and FBA, FFA, and VWFA are averaged across their sub-areas. Higher-level visual regions show greater interpretability than early visual areas—likely because they encode higher-level semantics that are both easier to decompose and to describe using natural language within our pipeline.

\begin{table*}[t]
\centering
\small
\setlength{\tabcolsep}{2pt}
\renewcommand{\arraystretch}{0.95}
\caption{\textbf{Per-ROI quantitative results.} Per-ROI percentage of explained hypotheses ($>0.5$) on the pool enriched with all predicted fMRI (+120k predicted fMRI). V1–V3 are averaged across ventral/dorsal; FBA, FFA, and VWFA are averages of their sub-areas.}
\begin{tabular}{lcccccccccccccc}
\toprule
Method & V1 & V2 & V3 & hV4 & EBA & OFA & OPA & OWFA & PPA & FFA & RSC & FBA & VWFA \\
\midrule
Voxels         & 0.2\% & 0.5\% & 0.6\% & 1.2\% & 5.8\% & 1.1\% & 2.8\% & 3.1\% & 2.1\% & 2.7\% & 1.0\% & 2.8\% & 3.0\% \\
PCA (Single)   & 0.3\% & 0.5\% & 0.5\% & 0.5\% & 4.4\% & 0.7\% & 2.3\% & 2.7\% & 2.9\% & 2.0\% & 1.8\% & 2.6\% & 3.2\% \\
NMF (Single)   & 0.2\% & 0.1\% & 0.4\% & 0.4\% & 2.0\% & 0.8\% & 1.1\% & 1.2\% & 1.7\% & 1.6\% & 2.2\% & 2.4\% & 1.1\% \\
ICA (Single)   & 1.9\% & 2.7\% & 5.7\% & 11.4\% & 15.3\% & 7.1\% & 13.9\% & 14.5\% & 12.8\% & 8.2\% & 6.2\% & 9.7\% & 0.2\% \\
SAE (Single)   & 0.6\% & 0.9\% & 1.3\% & 4.8\% & 14.5\% & 1.9\% & 9.2\% & 10.1\% & 6.8\% & 5.7\% & 2.9\% & 7.6\% & 7.6\% \\
SAE+ICA & 1.8\% & 2.8\% & 6.8\% & 11.0\% & 18.8\% & 7.5\% & 16.8\% & 17.2\% & 13.6\% & 8.6\% & 5.2\% & 10.7\% & 11.8\% \\
\bottomrule
\end{tabular}
\end{table*}

\paragraph{Complementarity across decompositions.}
We aim to understand which decomposition methods are most beneficial to combine, i.e., which ones provide complementary patterns that explain hypotheses the others do not. To this end, we measure, for every pair of decompositions, the change in the percentage of explained hypotheses (threshold 0.5) when combining their patterns. “Combining” a pair means pooling the patterns from both methods and recomputing the fraction of explained hypotheses; for reference, combining a method with itself corresponds to using multiple runs (different seeds) of the same decomposition. The reported value is the difference between the combined score and the maximum score of the two methods when used separately. As shown in \cref{tab:pairwise_methods}, SAE and ICA are the most complementary pair, each substantially improving the other.

\begin{table}[h]
\centering
\setlength{\tabcolsep}{8pt}
\caption{Pairwise complementarity between decompositions. Values show the gain in explained hypotheses (threshold 0.5) when combining each pair, relative to the better single method.}
\vspace{0.15cm}
\label{tab:pairwise_methods}
\begin{tabular}{@{}lccccc@{}}
\toprule
       & Voxels & PCA & NMF & ICA & SAE \\
\midrule
Voxels & 0.0\% & 0.5\% & 1.4\% & 0.1\% & -0.2\% \\
PCA    & 0.5\% & 0.0\% & 0.0\% & 0.4\% & 0.0\% \\
NMF    & 1.4\% & 0.0\% & 0.0\% & 0.6\% & 0.1\% \\
ICA    & 0.1\% & 0.4\% & 0.5\% & 0.2\% & 2.1\% \\
SAE    & -0.2\%& 0.0\% & 0.1\% & 2.1\% & 0.7\% \\
\bottomrule
\end{tabular}
\end{table}

\clearpage

\section{Extended Neuroscientific Validations}
\label{sec_sup:extended_validation}

\subsection{Alignment between ROI-discovered representations and known functionality}

In our method, we can ask, for each ROI, which concepts are represented in it, meaning which concepts have a brain pattern in this ROI with a high interpretability score. In parallel, we can also ask which concepts from our hypothesis dictionary align with the ROI's known functionality. We can then measure, for each ROI, what fraction of these functionally aligned concepts are indeed found to be interpretable by our method.

To create the ROI--concept alignment list, we use a strong LLM (\texttt{GPT-5-mini}\footnote{\url{https://developers.openai.com/api/docs/models/gpt-5-mini}}). We first ask it to describe, for each ROI, the general known functionality according to neuroscientific knowledge, considering only clear and widely accepted functional roles. Then, for each ROI and concept in our dictionary, we ask whether the concept is well aligned with the functionality of that ROI. The model is instructed to assign a concept to an ROI only when the concept clearly matches the ROI's functional category.
For example, the known functionality of FFA is faces, so both general concepts such as ``face'' or ``human face'' are included in the relevant concept list for this ROI, as well as more nuanced ones such as ``smiling face''. EBA is known to correspond to body parts and actions, so concepts such as ``hands,'' ``running,'' and ``skateboarding'' are all aligned as relevant. Concepts that do not clearly correspond to a known functional domain are not assigned to any ROI. This procedure provides, for each ROI, a set of \emph{functionally relevant} concepts, i.e., concepts that are assigned to that ROI by the LLM based on known functional selectivity.

\begin{figure*}[ht]
    \centering
    \vspace{-0.2cm}
    \includegraphics[width=0.59\linewidth]{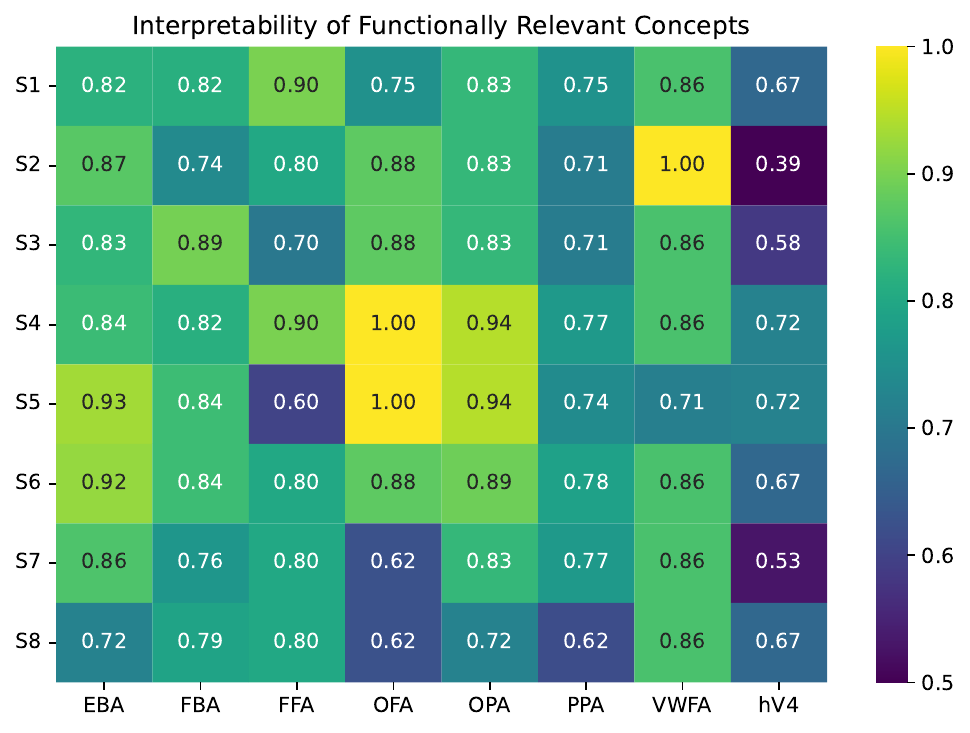}
    \vspace{-0.2cm}
    \caption{\textbf{Alignment between ROI-discovered representations and known functionality.}  For each subject (rows) and ROI (columns), we report the proportion of concepts labeled as functionally relevant to that ROI (based on automatic LLM assignment) that are also found to be represented in that ROI (score $> 0.5$). Higher values indicate stronger agreement between the known functional role of an ROI and the concepts identified by our method.}
    \label{fig:concept_roi_alignment}
\end{figure*}

We then use the score produced by our pipeline for each ROI and concept, namely the alignment score of the best-matching pattern in that ROI for that concept. We examine the relevant concept list for each ROI and measure what fraction of those concepts receives a score greater than 0.5 from our pipeline.
To ensure robustness across subjects, we restrict the analysis to \emph{globally interpretable concepts}, defined as concepts with interpretability score $> 0.5$ in at least three subjects (regardless of ROI). For each subject and ROI, we then define the set of \emph{interpretable concepts} as those with score $> 0.5$ in that specific subject and ROI. Both the relevant and interpretable sets are restricted to the global concept set. Finally, for each subject and ROI, we compute the alignment score as the fraction of relevant concepts (as defined above) that are also interpretable. ROIs with fewer than five relevant concepts are excluded from the analysis.
Results (Fig. ~\ref{fig:concept_roi_alignment}) show consistently high values across subjects and ROIs, indicating that concepts identified as interpretable within an ROI tend to match its known functional specialization.

\clearpage
\subsection{Cross-subject consistency of interpretable concepts}

To assess the robustness of discovered concepts across subjects, we measure the overlap of interpretable concepts identified in each ROI using an intersection-over-union (IoU) metric. For each ROI, we consider subsets of $k$ subjects and compute the IoU of their corresponding concept sets, defined as the size of the intersection divided by the size of the union. We report the average IoU across subsets of size $k=2,4,8$ (Table ~\ref{tab:Concept consistency across subjects.} in the main text).

To evaluate whether the observed overlap exceeds chance, we compare against a permutation-based null model, where for each subset we randomly assign ROIs to subjects and recompute the IoU. We use this null distribution to assess statistical significance.

\subsection{Cross-subject consistency of interpretable patterns}
\label{sec_sup:cross_subject_pattern}

To further validate the biological relevance of the discovered SAE patterns, we asked whether patterns found in one ROI tend to have similar counterparts in the same or functionally related ROIs, both within and across subjects. For this analysis, we considered each SAE pattern in Subject 1 and measured its overlap with patterns from other ROIs using the intersection-over-union (IoU) between their sets of top-activating images. For every pattern in Subject 1, and for each target ROI separately, we identified the pattern with the highest IoU in that ROI. We then averaged these best-match scores across all patterns belonging to a given ROI in Subject 1, producing an ROI-to-ROI similarity matrix.

We visualize these results as heatmaps. Since within-Subject-1 comparisons contain especially high values on the diagonal, reflecting similarity between different SAE patterns learned in the same ROI, we do not show the Subject-1-to-Subject-1 diagonal.

Although the absolute IoU values are modest, since having exactly the same top-activating images is unlikely, the resulting structure is informative. First, the heatmaps are broadly consistent across subjects, suggesting that the discovered SAE patterns capture reproducible aspects of functional organization. Second, ROIs with similar functionality tend to show higher mutual similarity. In particular, early visual ventral regions cluster more strongly with each other than with dorsal regions, and additional ROIs with related functionality also exhibit stronger overlap. This indicates that the SAE-derived patterns are not only interpretable in terms of fine-grained visual concepts, but also 
capture the broader functional organization of the cortex, both within individual subjects and across subjects.

\begin{figure*}[ht]
  \centering
\includegraphics[width=0.83\linewidth]{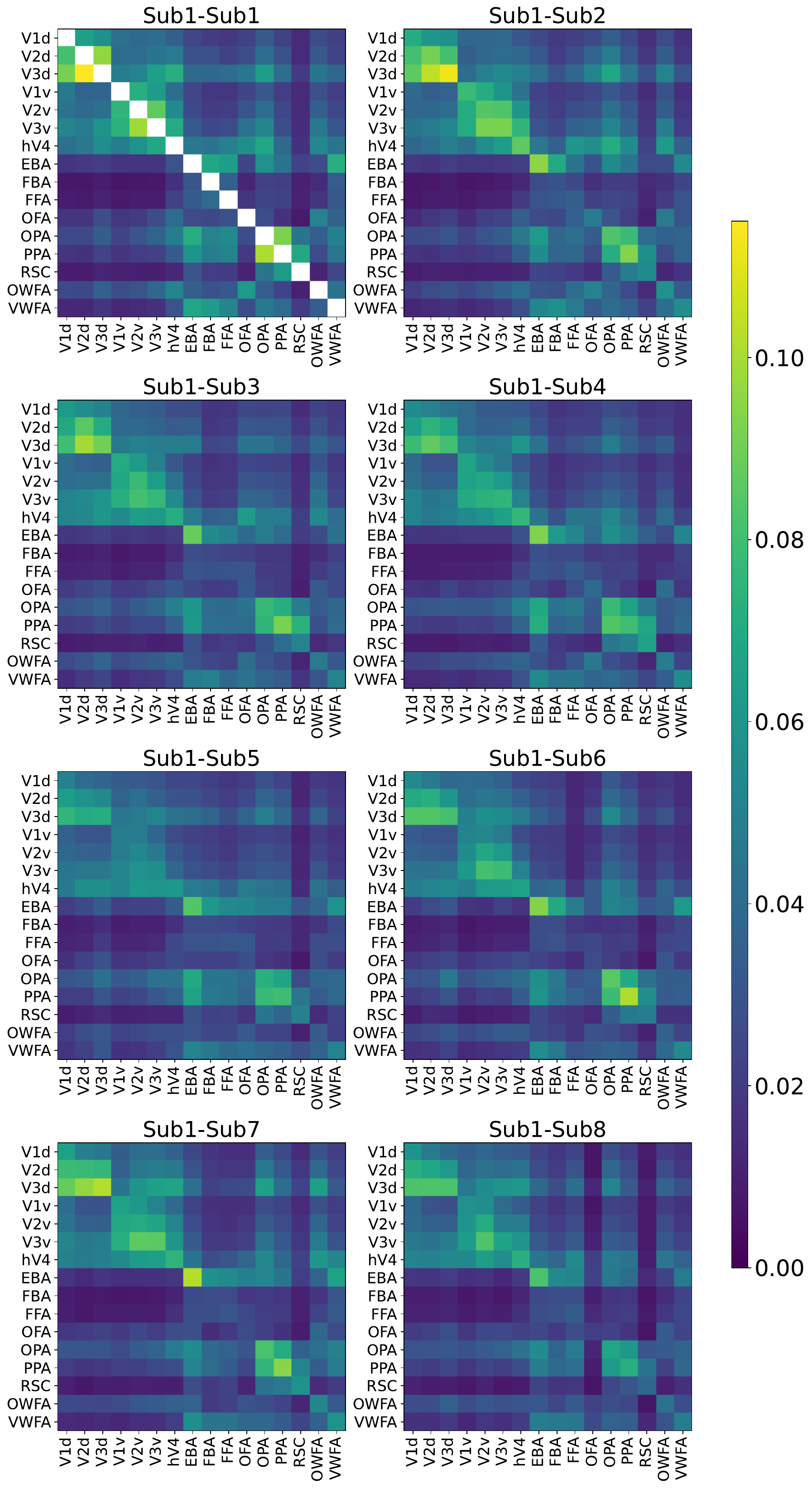}
\vspace{-0.1cm}
  \caption{\textbf{Cross-subject consistency of SAE patterns.} For each SAE pattern in Subject 1, we find the pattern with the highest IoU of top-activating images in each target ROI, within the same subject and across other subjects, and average these scores across patterns from each source ROI. Functionally related ROIs show higher similarity, and the overall structure is consistent across subjects.}
  \label{fig:IoU_across_subjects}
\end{figure*}

\clearpage
\section{Additional Details} 
\label{sec_sup:additional_details}

\vspace{-0.2cm}
\subsection{SAE Training Details}
\label{sec_sup:SAE_training}
\vspace{-0.1cm}

The Sparse Autoencoder (SAE) was trained to learn a compact and interpretable latent representation of fMRI activations, enabling decomposition into meaningful spatial patterns.

\vspace{-0.3cm}
\paragraph{Architecture.}
We used a simple, fully linear autoencoder, where both encoder and decoder are implemented as matrix multiplications. A ReLU activation is applied after the encoder to enforce non-negativity in the latent activations. Sparsity is encouraged via an $\ell_1$ loss applied to the latent space, added to the reconstruction mean-squared error (MSE) loss. The \textit{expansion factor} controls the latent dimensionality (the ratio between latent and input size), while the \textit{sparsity coefficient} balances the sparsity loss relative to the reconstruction loss.

\vspace{-0.3cm}
\paragraph{Combining with predicted fMRI.}
We combine the real measured fMRI data ($\sim$10k samples) with predicted fMRI activations from the image-to-fMRI encoder ($\sim$120k samples). Real fMRI measurements are inherently noisy, and many voxels exhibit low signal-to-noise ratio (SNR). Consequently, the predicted encoder often outputs values close to zero for those voxels, leading to a mismatch in distribution between measured and predicted fMRI. Training a single SAE on this mixed data causes an imbalance, resulting in higher sparsity for predicted samples and reduced sparsity for measured ones.  
To address this, we use separate encoder networks for the measured and predicted data, while sharing a common decoder. This allows both data types to be decomposed into the same set of components (dictionary patterns) while enabling independent control over their activations. During training, each batch contained an equal number of measured and predicted samples to ensure balanced gradients from both data sources.

\vspace{-0.3cm}
\paragraph{Hyperparameters.}
As reported in the ablation studies, we experimented with multiple combinations of sparsity coefficients and expansion factors. For the main hyperparameter search, we considered values $\{1,2,4\}$ for both. The top four models selected according to the rank score threshold of 0.8 on Subject~1, were used in the \textit{multiple model} setup: sparsity coefficient~4 with expansion factors $\{1,2,4\}$ and sparsity coefficient~1 with expansion factor~2.  
The \textit{single model} configuration used sparsity coefficient~4 and expansion factor~4. The same hyperparameters were applied across all subjects.

\vspace{-0.15cm}
\subsection{Baseline Decomposition Training Details}
\label{sec_sup:decompostions_baselines}
\vspace{-0.2cm}

All baseline decomposition methods were trained in two configurations: (i) using only the measured fMRI data ($\sim$10k samples), and (ii) using the combined pool of measured and predicted fMRI activations ($\sim$120k samples). This setup enables consistent comparison between the baselines and the Sparse Autoencoder (SAE), both when relying solely on measured data and when leveraging the large scale predicted data.

\vspace{-0.35cm}
\paragraph{Voxels.}
The voxel baseline uses the raw voxel activations without any decomposition or learned transformation. Each voxel is treated as an independent component. For visualization and interpretability, every voxel’s activation pattern was duplicated into two components (the original and its negation), allowing retrieval of top activating images for both positive and negative responses. This ensures interpretability is consistent across all decomposition methods, including those that produce only nonnegative activations.

\vspace{-0.35cm}
\paragraph{PCA.}
Principal Component Analysis (PCA) serves as a deterministic, orthogonal decomposition baseline. It was trained on both measured only and combined fMRI datasets, with the number of components determined by the cumulative explained variance.  
To ensure consistency for interpretability visualization, each PCA component was duplicated into two versions, positive and negative, allowing activation maps to always correspond to positively responsive image sets. 

\vspace{-0.35cm}
\paragraph{ICA.}
Independent Component Analysis (ICA) was applied using numbers of components corresponding to the dimensionalities that achieve 90\%, 95\%, and 98\% explained variance in PCA. Higher explained variance targets ($>$98\%) were found to be unstable or unnecessary due to overfitting. ICA was run multiple times with different random seeds to capture variability arising from its nondeterministic optimization process. Similar to PCA, each independent component was duplicated into its positive and negative versions for interpretability visualization.

\vspace{-0.35cm}
\paragraph{NMF.}
Nonnegative Matrix Factorization (NMF) was trained with the same component counts as ICA (matching the 0.9, 0.95, and 0.98 explained variance configurations). Since NMF produces strictly nonnegative components, no positive or negative duplication was needed. NMF was also trained on both measured and combined fMRI data. Different hyperparameter settings (for example, initialization and regularization) showed minimal impact on the final interpretability results, consistent with its stable convergence properties under nonnegativity constraints.

\vspace{-0.15cm}
\subsection{Visualize \& Explain}
\label{sec_sup:visual_explain}
\vspace{-0.1cm}

As described in the paper, in the \emph{Visualize} step we select, for every pattern, its top-\(N\) activating responses and collect their corresponding images, producing a set of images that \emph{visualize} what most strongly activates that pattern. This is done separately for measured fMRI responses and for predicted responses. We take the six most activating images from the measured fMRI pool and the ten most activating images from the predicted fMRI pool (16 images in total).  

For each image, we first generate a detailed caption using \texttt{Qwen2.5-VLM-32B}\footnote{\url{https://huggingface.co/Qwen/Qwen2.5-VL-32B-Instruct}}. The VLM is instructed (see prompt in \cref{sup_fig:image_captioning_prompt}) to produce rich descriptions including objects, scene or room type, dominant colors, ongoing activities, and body postures.  

Given the set of \(N\) captions, we then ask a language model  \texttt{Qwen2.5-32B}\footnote{\url{https://huggingface.co/Qwen/Qwen2.5-32B-Instruct}} to generate between 3 and 12 hypotheses that may explain what is shared among at least half of those images (\cref{sup_fig:Hyp_gen_prompt}). Requesting multiple hypotheses provides diversity, since both the captions and the LLM can be biased toward a dominant concept, whereas smaller but meaningful shared attributes might otherwise be missed.  

We intentionally separate the VLM and LLM stages. We found that current vision–language models often struggle to infer what is shared across multiple images, frequently producing incorrect or overly general hypotheses. By contrast, separating the process allows each VLM to focus on a single image, producing high-quality detailed captions, while the LLM operates only on textual descriptions—an easier and more reliable input for identifying cross-image commonalities.

\vspace{-0.1cm}
\subsection{Scaling to an Unlimited Number of Patterns}
\label{sec_sup:scaling}
\vspace{-0.1cm}

Explaining every component in every ROI and for every decomposition method is costly and inefficient. To scale interpretability, we introduce two steps: 
(i) \textbf{\emph{Hypothesis dictionary generation}}, and (ii) \textbf{\emph{Hypothesis–image labeling}}. Once the dictionary and the image labels are in place, any new brain pattern can be evaluated by measuring how consistently its top-activating images express concepts from the dictionary. The prompts used for hypothesis–image labeling are shown in \cref{sup_fig:Hyp_label_prompt}, and examples of images with positively labeled hypotheses appear in \cref{sup_fig:image_labeled}.

\subsection{Region of Interest (ROI)}
\label{sec_sup:ROI}

We analyze activity within predefined Regions of Interest (ROIs), which are standard cortical parcels used to summarize responses within broader anatomical and functional subdivisions of the visual system. In the Algonauts/NSD data, these ROIs are supplied by the dataset releases based on independent localizers and anatomical parcellations. The full set used here is:
V1v, V1d, V2v, V2d, V3v, V3d, hV4, VWFA-1, VWFA-2, EBA, FBA-1, FBA-2, FFA-1, FFA-2, OFA, OPA, OWFA, PPA, and RSC.
Early visual areas (V1–V3, separated into ventral and dorsal subregions) correspond to the low-level visual cortex and are primarily involved in processing basic visual attributes such as orientation, color, and spatial frequency. Area hV4 is associated with color and shape processing. Higher visual regions show more specialized selectivity: EBA, FBA, and OFA are body- and face-selective regions; OPA, PPA, and RSC respond to scenes, places, and navigation-related information; VWFA and OWFA are selective to word and object forms, respectively. Together, these ROIs span the visual hierarchy from low-level visual representations to high-level semantic and categorical processing.

\subsection{Compuational Resourences}
\label{sec:resources}
The hypothesis generation over image sets was performed using large LLM and VLM models for the upscaling stage on an H200 GPU, requiring around 100 GPU hours. The image labeling was performed on an L40S GPU using a smaller 8B VLM model, and required around 100 GPU hours. After the upscaling stage, the pipeline itself was run on L40S GPUs; the full pipeline over all decompositions and concepts took around 4 hours.

\clearpage
\section{Additional Quantitative Evaluation}
\label{sec_sup:additional_Quantitative}
\vspace{-0.15cm}

\begin{table}[ht]
\centering
\setlength{\tabcolsep}{10pt}
\small
  \centering
  \caption{\textbf{Number of Interpretable Patterns.} The number of patterns whose best hypothesis alignment exceeds the same threshold (0.5 or 0.8). To avoid counting near-duplicate patterns, we remove any component whose voxel-wise correlation with a higher-scoring pattern exceeds 0.5}
  \vspace{0.1cm}
  \begin{tabular}{@{}l*{4}{c}@{}}
  \toprule
  & \multicolumn{2}{c}{Measured fMRI}
  & \multicolumn{2}{c}{+ Predicted fMRI} \\
  \cmidrule(lr){2-3}\cmidrule(lr){4-5}
  Method & $>0.5$ & $>0.8$ & $>0.5$ & $>0.8$ \\
  \midrule
  Voxels         & 5234 & 355 & 5905 & 291 \\
  PCA (Single)           & 1045 & 6 & 76 & 6 \\
  NMF (Single)            & 27 & 2 & 19 & 2 \\
  \addlinespace[2pt]
  ICA (Single)   & 580 & 4 & 226 & 44 \\
  ICA (Multiple) & 424 & 41 & 305 & 61 \\
  \addlinespace[2pt]
  SAE (Single)   & 17242 & 586 & 8858 & 286 \\
  SAE (Multiple) & 30005 & 1074 & 15748 & 617 \\
  \addlinespace[2pt]
  SAE+ICA        & 30583 & 1077 & 16051 & 679 \\
  \bottomrule
  \label{tab:patterns}
  \vspace{-0.5cm}
  \end{tabular}
\end{table}

\begin{table}[h]
\centering
\setlength{\tabcolsep}{6pt}
\small
\caption{\textbf{Average across subjects: Interpretable Hypotheses (count) with standard error.}}
\vspace{0.15cm}
\begin{tabular}{@{}l*{4}{c}@{}}
\toprule
& \multicolumn{2}{c}{Measured fMRI}
& \multicolumn{2}{c}{+ Predicted fMRI} \\
\cmidrule(lr){2-3}\cmidrule(lr){4-5}
Method & $>0.5$ & $>0.8$ & $>0.5$ & $>0.8$ \\
\midrule
Voxels
& $42.1 \pm 4.9$
& $4.5 \pm 0.8$
& $86.5 \pm 7.0$
& $15.9 \pm 1.9$ \\
PCA
& $20.4 \pm 4.8$
& $4.2 \pm 1.5$
& $78.5 \pm 7.0$
& $11.1 \pm 2.1$ \\
NMF
& $19.2 \pm 4.2$
& $3.0 \pm 0.9$
& $61.5 \pm 9.0$
& $10.4 \pm 2.2$ \\
\addlinespace[2pt]
ICA
& $11.8 \pm 3.7$
& $2.9 \pm 1.5$
& $207.0 \pm 19.8$
& $34.9 \pm 9.4$ \\
\addlinespace[2pt]
SAE
& $104.6 \pm 10.5$
& $17.8 \pm 4.0$
& $218.9 \pm 14.0$
& $50.5 \pm 6.3$ \\
\addlinespace[2pt]
SAE+ICA
& $104.5 \pm 10.4$
& $17.9 \pm 4.1$
& $250.1 \pm 18.2$
& $58.0 \pm 9.7$ \\
\bottomrule
\end{tabular}
\label{tab_sup:hyp_Avg_nubmer}

\end{table}

\begin{table}[h]
\centering
\setlength{\tabcolsep}{6pt}
\small
\caption{\textbf{Average across subjects: Interpretable Hypotheses (percentage) with standard error.}}
\vspace{0.15cm}
\begin{tabular}{@{}l*{4}{c}@{}}
\toprule
& \multicolumn{2}{c}{Measured fMRI}
& \multicolumn{2}{c}{+ Predicted fMRI} \\
\cmidrule(lr){2-3}\cmidrule(lr){4-5}
Method & $>0.5$ & $>0.8$ & $>0.5$ & $>0.8$ \\
\midrule
Voxels
& $3.17 \pm 0.37\%$
& $0.34 \pm 0.06\%$
& $6.50 \pm 0.52\%$
& $1.19 \pm 0.15\%$ \\
PCA
& $1.53 \pm 0.36\%$
& $0.32 \pm 0.12\%$
& $5.90 \pm 0.53\%$
& $0.84 \pm 0.16\%$ \\
NMF
& $1.45 \pm 0.32\%$
& $0.23 \pm 0.07\%$
& $4.62 \pm 0.68\%$
& $0.78 \pm 0.16\%$ \\
\addlinespace[2pt]
ICA
& $0.88 \pm 0.28\%$
& $0.22 \pm 0.11\%$
& $15.56 \pm 1.49\%$
& $2.62 \pm 0.71\%$ \\
\addlinespace[2pt]
SAE
& $7.87 \pm 0.79\%$
& $1.33 \pm 0.30\%$
& $16.46 \pm 1.05\%$
& $3.80 \pm 0.48\%$ \\
\addlinespace[2pt]
SAE+ICA
& $7.86 \pm 0.78\%$
& $1.34 \pm 0.31\%$
& $18.81 \pm 1.37\%$
& $4.36 \pm 0.73\%$ \\
\bottomrule
\end{tabular}
\label{tab_sup:hyp_Avg_precentage}
\end{table}

\begin{table*}[t]
\centering
\small
\setlength{\tabcolsep}{6pt}
\caption{\textbf{Interpretable hypotheses across all subjects (count / percentage).}}
\begin{tabular}{llcccc}
\toprule
\textbf{Subject} & \textbf{Method} & \multicolumn{2}{c}{Measured fMRI} & \multicolumn{2}{c}{+ Predicted fMRI} \\
\cmidrule(lr){3-4}\cmidrule(lr){5-6}
& & >0.5 & >0.8 & >0.5 & >0.8 \\
\midrule

\multirow{6}{*}{S1}
& Voxels  & 39 / 2.9\% & 5 / 0.4\% & 88 / 6.6\% & 18 / 1.4\% \\
& PCA     & 32 / 2.4\% & 11 / 0.8\% & 97 / 7.3\% & 13 / 1.0\% \\
& NMF     & 18 / 1.4\% & 3 / 0.2\% & 73 / 5.5\% & 10 / 0.8\% \\
& ICA     & 11 / 0.8\% & 0 / 0.0\% & 244 / 18.3\% & 72 / 5.4\% \\
& SAE     & 82 / 6.2\% & 17 / 1.3\% & 231 / 17.4\% & 51 / 3.8\% \\
& SAE+ICA & 83 / 6.2\% & 16 / 1.2\% & 286 / 21.5\% & 78 / 5.9\% \\
\midrule

\multirow{6}{*}{S2}
& Voxels  & 54 / 4.1\% & 5 / 0.4\% & 107 / 8.0\% & 19 / 1.4\% \\
& PCA     & 30 / 2.3\% & 5 / 0.4\% & 96 / 7.2\% & 18 / 1.4\% \\
& NMF     & 28 / 2.1\% & 5 / 0.4\% & 84 / 6.3\% & 17 / 1.3\% \\
& ICA     & 15 / 1.1\% & 2 / 0.2\% & 282 / 21.2\% & 48 / 3.6\% \\
& SAE     & 115 / 8.6\% & 13 / 1.0\% & 247 / 18.6\% & 61 / 4.6\% \\
& SAE+ICA & 115 / 8.6\% & 13 / 1.0\% & 310 / 23.3\% & 70 / 5.3\% \\
\midrule

\multirow{6}{*}{S3}
& Voxels  & 31 / 2.3\% & 4 / 0.3\% & 72 / 5.4\% & 13 / 1.0\% \\
& PCA     & 17 / 1.3\% & 2 / 0.2\% & 72 / 5.4\% & 8 / 0.6\% \\
& NMF     & 14 / 1.1\% & 1 / 0.1\% & 44 / 3.3\% & 5 / 0.4\% \\
& ICA     & 5 / 0.4\% & 1 / 0.1\% & 175 / 13.1\% & 14 / 1.1\% \\
& SAE     & 82 / 6.2\% & 7 / 0.5\% & 191 / 14.4\% & 46 / 3.5\% \\
& SAE+ICA & 82 / 6.2\% & 7 / 0.5\% & 218 / 16.4\% & 45 / 3.4\% \\
\midrule

\multirow{6}{*}{S4}
& Voxels  & 41 / 3.1\% & 4 / 0.3\% & 83 / 6.2\% & 13 / 1.0\% \\
& PCA     & 20 / 1.5\% & 2 / 0.2\% & 60 / 4.5\% & 9 / 0.7\% \\
& NMF     & 20 / 1.5\% & 2 / 0.2\% & 36 / 2.7\% & 6 / 0.5\% \\
& ICA     & 7 / 0.5\% & 2 / 0.2\% & 168 / 12.6\% & 31 / 2.3\% \\
& SAE     & 103 / 7.7\% & 15 / 1.1\% & 189 / 14.2\% & 40 / 3.0\% \\
& SAE+ICA & 103 / 7.7\% & 16 / 1.2\% & 202 / 15.2\% & 42 / 3.2\% \\
\midrule

\multirow{6}{*}{S5}
& Voxels  & 71 / 5.3\% & 4 / 0.3\% & 122 / 9.2\% & 25 / 1.9\% \\
& PCA     & 43 / 3.2\% & 11 / 0.8\% & 109 / 8.2\% & 17 / 1.3\% \\
& NMF     & 43 / 3.2\% & 7 / 0.5\% & 107 / 8.0\% & 21 / 1.6\% \\
& ICA     & 36 / 2.7\% & 13 / 1.0\% & 269 / 20.2\% & 74 / 5.6\% \\
& SAE     & 159 / 11.9\% & 42 / 3.2\% & 262 / 19.7\% & 85 / 6.4\% \\
& SAE+ICA & 159 / 11.9\% & 43 / 3.2\% & 303 / 22.8\% & 110 / 8.3\% \\
\midrule

\multirow{6}{*}{S6}
& Voxels  & 30 / 2.3\% & 4 / 0.3\% & 66 / 5.0\% & 17 / 1.3\% \\
& PCA     & 8 / 0.6\% & 1 / 0.1\% & 66 / 5.0\% & 4 / 0.3\% \\
& NMF     & 15 / 1.1\% & 5 / 0.4\% & 59 / 4.4\% & 5 / 0.4\% \\
& ICA     & 6 / 0.5\% & 1 / 0.1\% & 210 / 15.8\% & 14 / 1.1\% \\
& SAE     & 94 / 7.1\% & 12 / 0.9\% & 226 / 17.0\% & 36 / 2.7\% \\
& SAE+ICA & 93 / 7.0\% & 12 / 0.9\% & 251 / 18.9\% & 36 / 2.7\% \\
\midrule

\multirow{6}{*}{S7}
& Voxels  & 38 / 2.9\% & 9 / 0.7\% & 87 / 6.5\% & 16 / 1.2\% \\
& PCA     & 3 / 0.2\% & 1 / 0.1\% & 75 / 5.6\% & 17 / 1.3\% \\
& NMF     & 3 / 0.2\% & 0 / 0.0\% & 57 / 4.3\% & 13 / 1.0\% \\
& ICA     & 10 / 0.8\% & 3 / 0.2\% & 193 / 14.5\% & 19 / 1.4\% \\
& SAE     & 132 / 9.9\% & 26 / 2.0\% & 257 / 19.3\% & 58 / 4.4\% \\
& SAE+ICA & 131 / 9.8\% & 26 / 2.0\% & 266 / 20.0\% & 58 / 4.4\% \\
\midrule

\multirow{6}{*}{S8}
& Voxels  & 33 / 2.5\% & 1 / 0.1\% & 67 / 5.0\% & 6 / 0.5\% \\
& PCA     & 10 / 0.8\% & 1 / 0.1\% & 53 / 4.0\% & 3 / 0.2\% \\
& NMF     & 13 / 1.0\% & 1 / 0.1\% & 32 / 2.4\% & 6 / 0.5\% \\
& ICA     & 4 / 0.3\% & 1 / 0.1\% & 115 / 8.6\% & 7 / 0.5\% \\
& SAE     & 70 / 5.3\% & 10 / 0.8\% & 148 / 11.1\% & 27 / 2.0\% \\
& SAE+ICA & 70 / 5.3\% & 10 / 0.8\% & 165 / 12.4\% & 25 / 1.9\% \\
\bottomrule
\end{tabular}
\end{table*}

\clearpage
\section{Additional Visualizations}
\label{sec_sup:additional_Visualizations}

\begin{figure}[h]
  \centering
\includegraphics[width=0.80\linewidth]{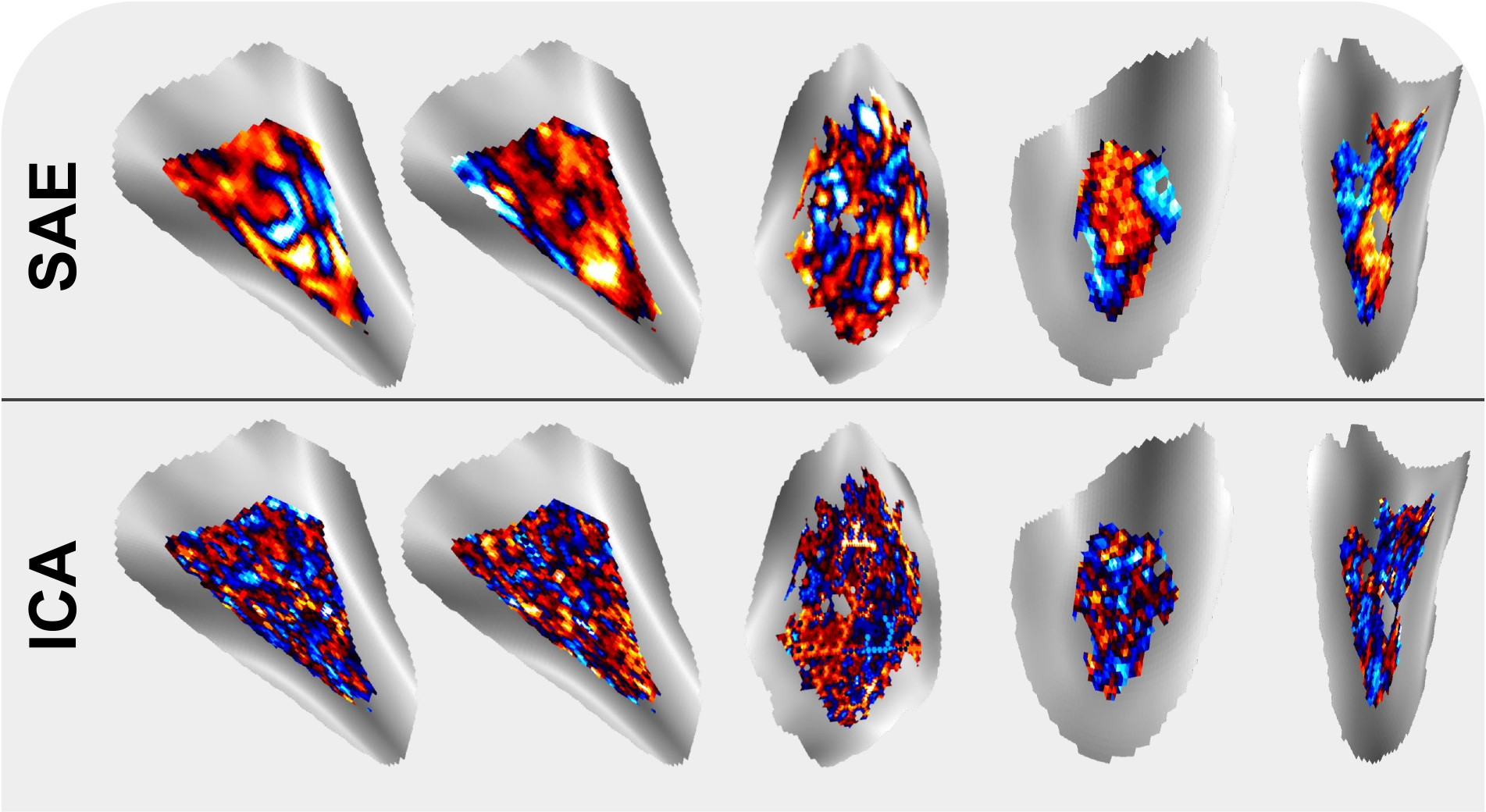}
  \caption{\textbf{Brain activation map (SAE vs ICA)}. We show different activation patterns derived from a Sparse Autoencoder (SAE) and Independent Component Analysis (ICA) for different ROIs. SAE patterns are notably more compact and clustered and spatially localized, despite both receiving only 1D voxel vectors and no spatial information.}
  \vspace{-0.25cm}
  \label{fig:SAE_vs_ICA_localization}
\end{figure}

\paragraph{Spatial localization of interpretable patterns (SAE vs.\ ICA).}
Extending the main-text observation that SAE patterns are more spatially localized, \cref{fig:SAE_vs_ICA_localization} presents side-by-side brain activation maps for multiple ROIs, comparing patterns derived from ICA and from SAE. Despite being trained on one-dimensional voxel vectors with no spatial priors, SAE produces compact, clustered patterns, whereas ICA frequently yields more diffuse or scattered maps.  All maps were generated using identical preprocessing, color scaling, and selection criteria (top-activating images per pattern), ensuring that visual differences reflect the decomposition method rather than visualization settings.

\begin{figure*}[ht]
  \centering
\includegraphics[width=\linewidth]{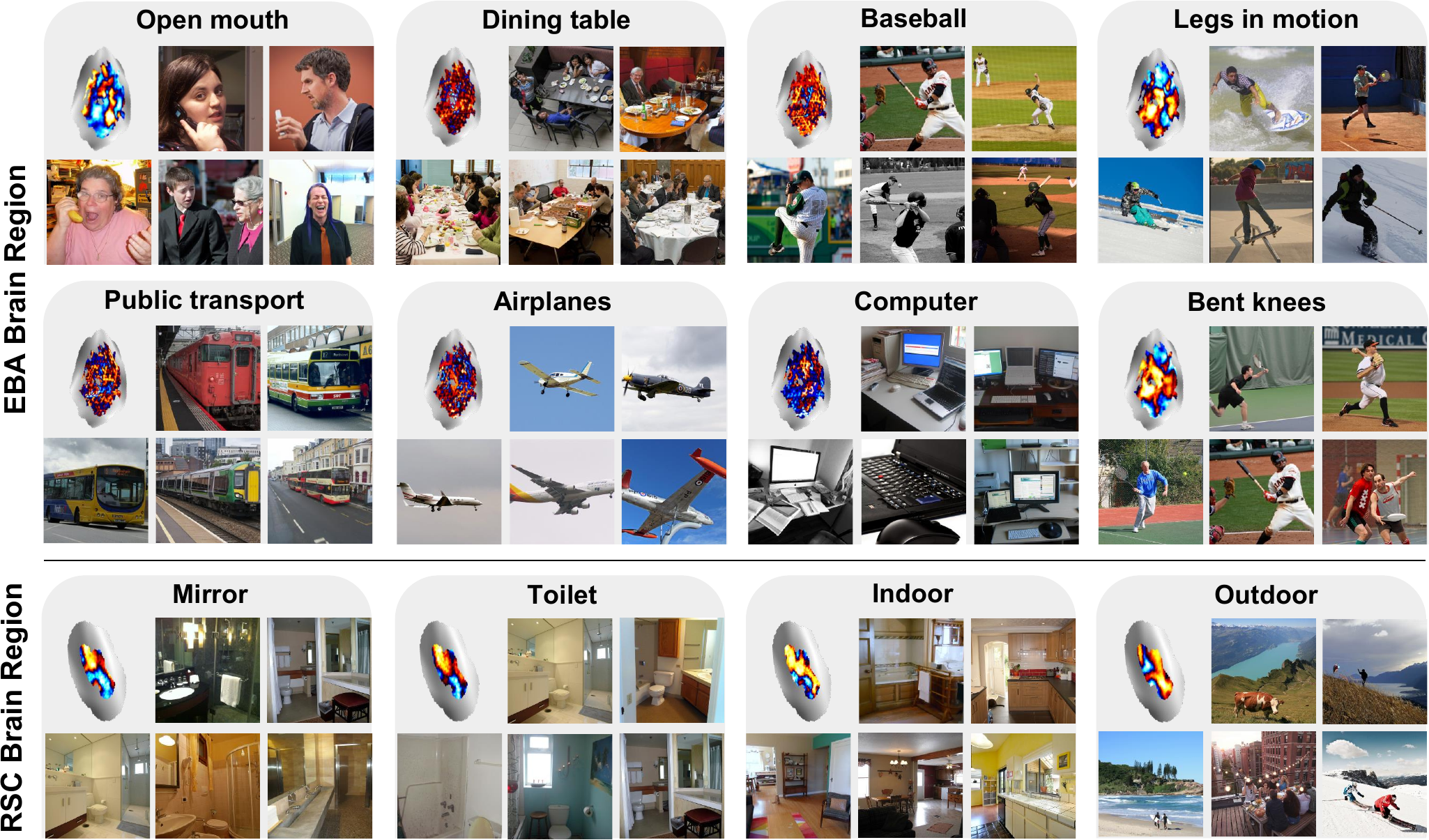}
  \caption{\textbf{Discovered Interpretable Patterns (EBA and RSC)}. We show additional patterns for Subject 1 with top activating images and selected explanations. EBA is known to encode bodies and actions, whereas RSC processes scene information.}  
  \label{fig:extra_concepts_1}
\end{figure*}

\begin{figure*}[ht]
  \centering
\includegraphics[width=\linewidth]{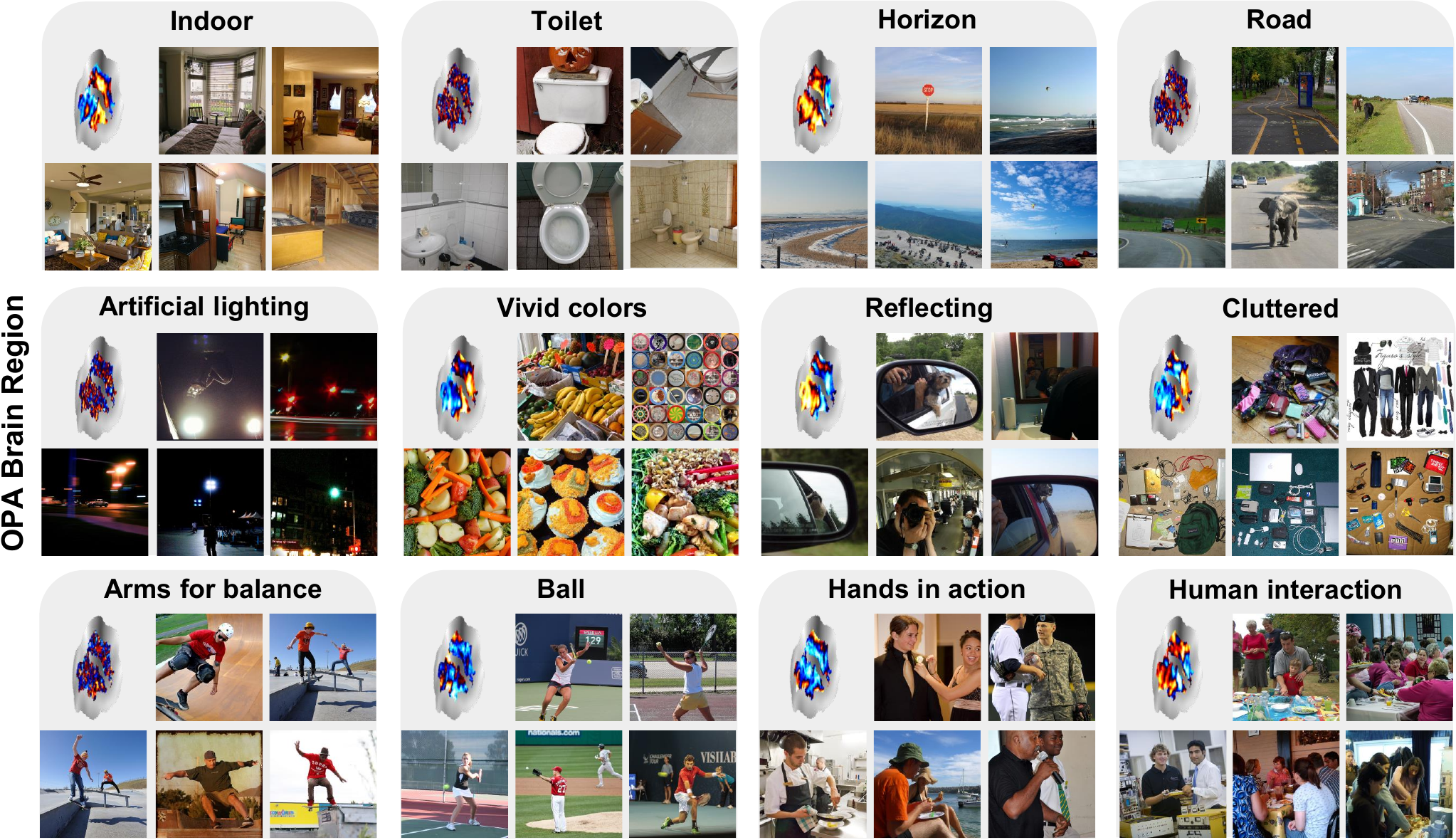}
  \caption{\textbf{Discovered Interpretable Patterns (OPA)}. We show additional patterns for Subject 1 with top activating images and selected explanations. OPA is known to process scene layout and navigability.}  
  \label{fig:extra_concepts_2}
\end{figure*}

\begin{figure*}[ht]
  \centering
\includegraphics[width=\linewidth]{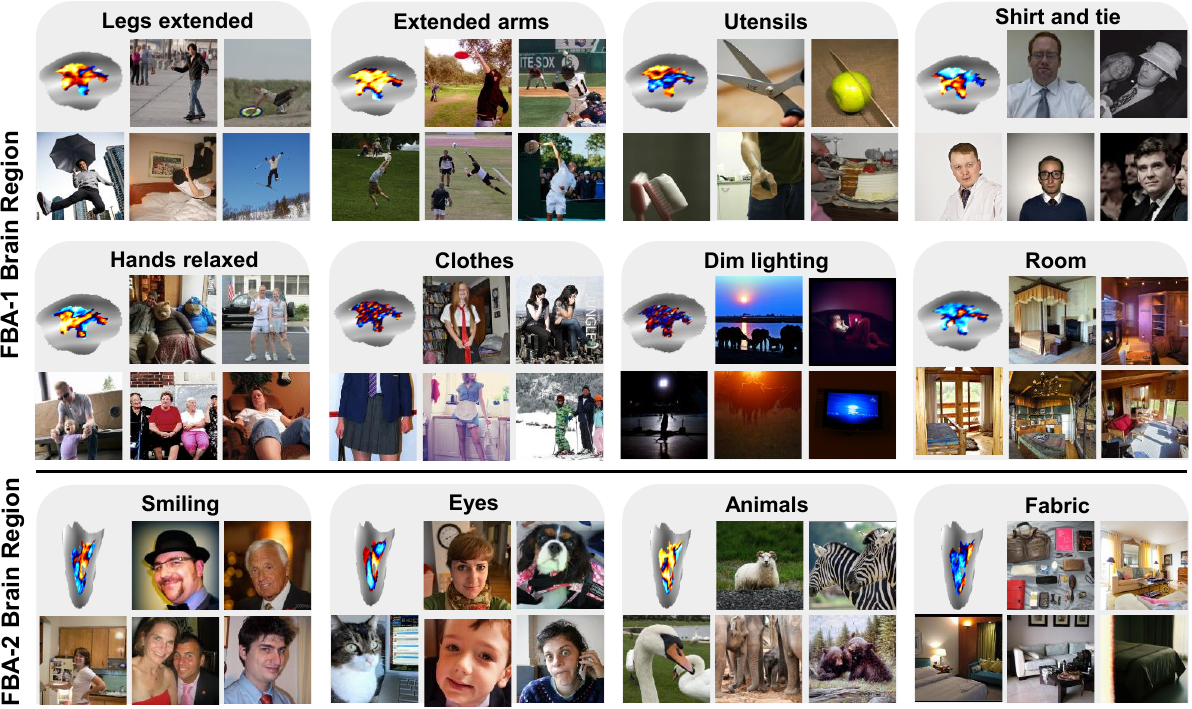}
  \caption{\textbf{Discovered Interpretable Patterns (FBA-1 and FBA-2)}. We show additional patterns for Subject 1 with top activating images and selected explanations. FBA is a body-selective area, encompassing two sub-areas shown here, FBA-1 and FBA-2.}  
  \label{fig:extra_concepts_3}
\end{figure*}

\vspace{-0.15cm}
\paragraph{Additional concepts across ROIs.}
We present additional interpretable patterns from Subject~1 together with representative top-activating images. We show further results for EBA and RSC in \cref{fig:extra_concepts_1}, for OPA in \cref{fig:extra_concepts_2}, and for FBA-1 and FBA-2 in \cref{fig:extra_concepts_3}.

\begin{figure*}[ht]
  \centering
\includegraphics[width=\linewidth]{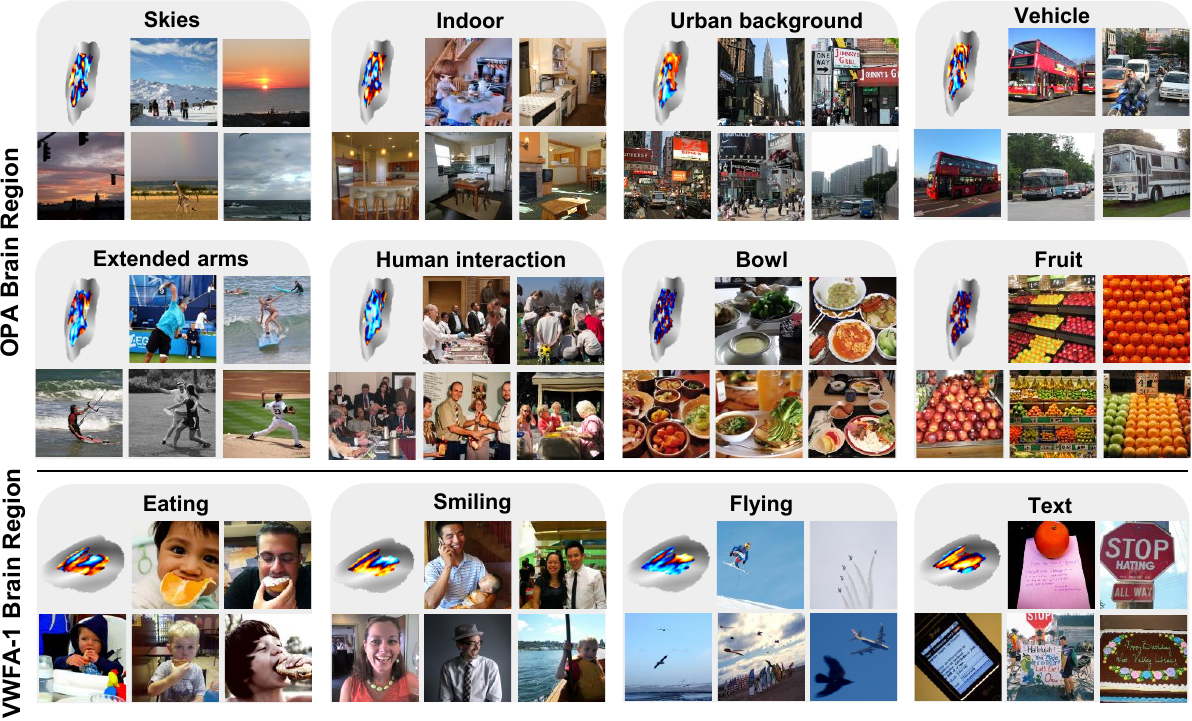}
  \caption{\textbf{Discovered Interpretable Patterns of Subject 2 (OPA and VWFA-1)}. We show patterns for Subject 2 with top activating images and selected explanations. OPA is known to process scene layout and navigability, and VWFA is involved in processing word shapes and visual text.}  
  \label{fig:sub2_concepts_1}
\end{figure*}

\begin{figure*}[ht]
  \centering
\includegraphics[width=\linewidth]{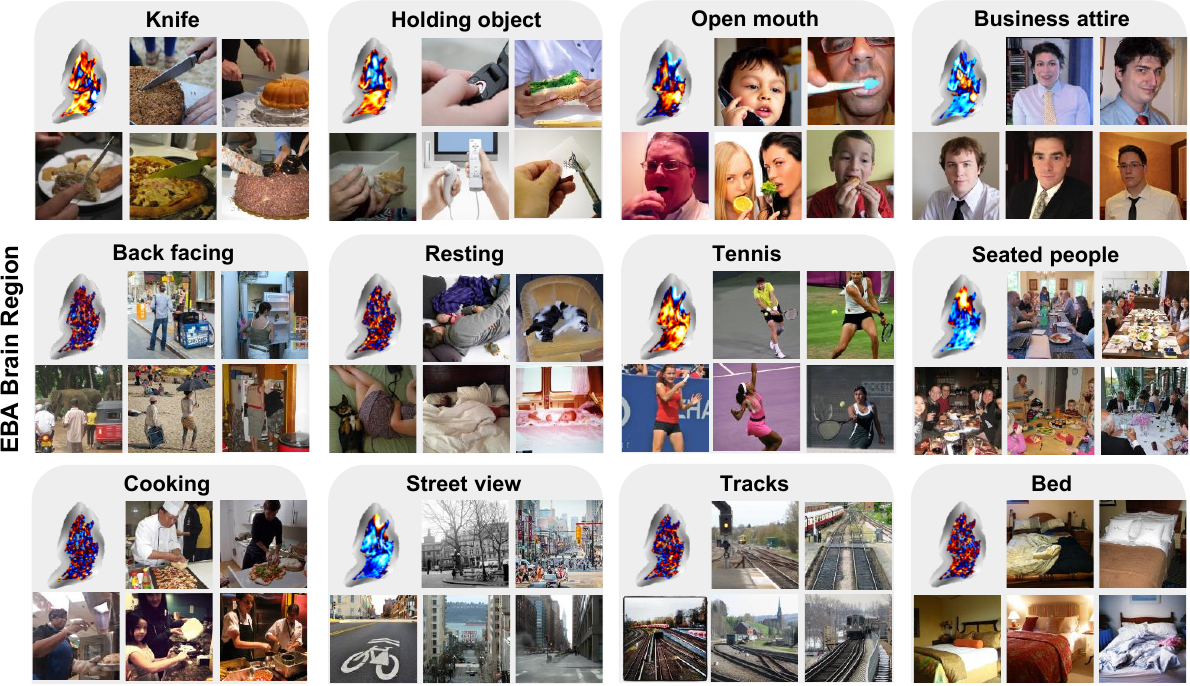}
  \caption{\textbf{Discovered Interpretable Patterns of Subject 2 (EBA)}. We show patterns for subject 2 with top activating images and selected explanations. EBA is known to encode bodies and actions.}  
  \label{fig:sub2_concepts_2}
\end{figure*}

\begin{figure*}[ht]
  \centering
\includegraphics[width=\linewidth]{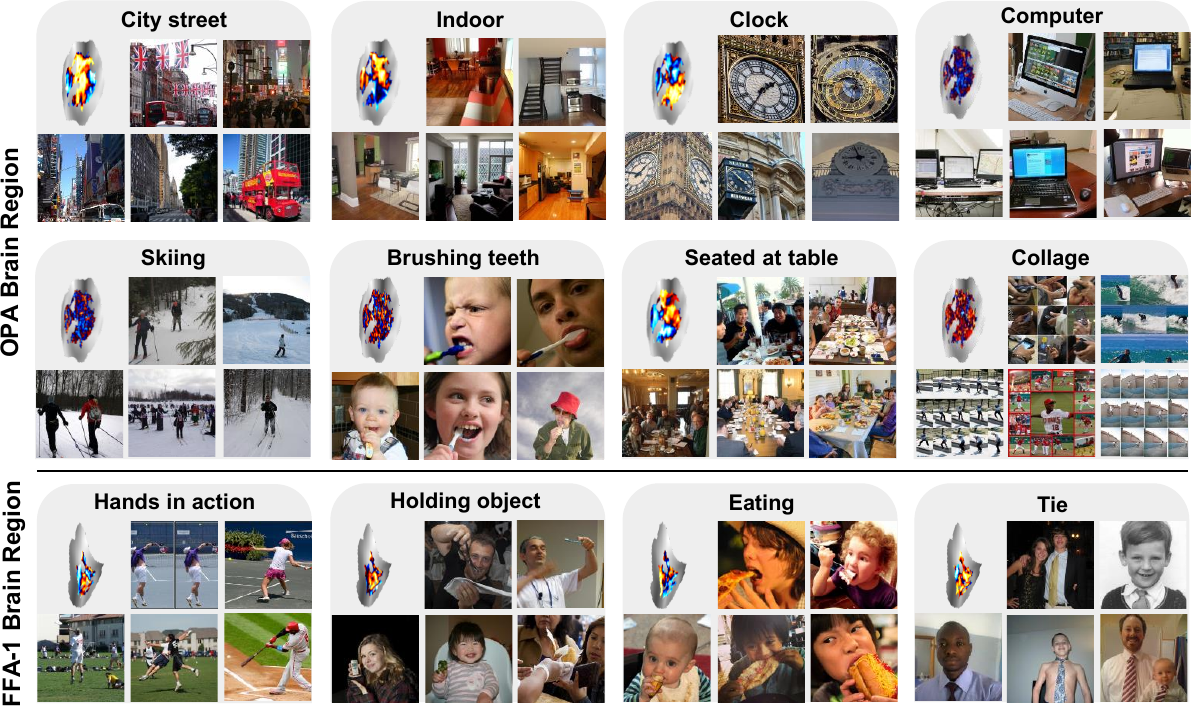}
  \caption{\textbf{Discovered Interpretable Patterns of Subject 5 (OPA and FFA-1)}. We show patterns for subject 5 with top activating images and selected explanations. OPA is known to process scene layout and navigability, and FFA is primarily known for face processing.}  
  \label{fig:sub5_concepts_1}
\end{figure*}

\begin{figure*}[ht]
  \centering
\includegraphics[width=\linewidth]{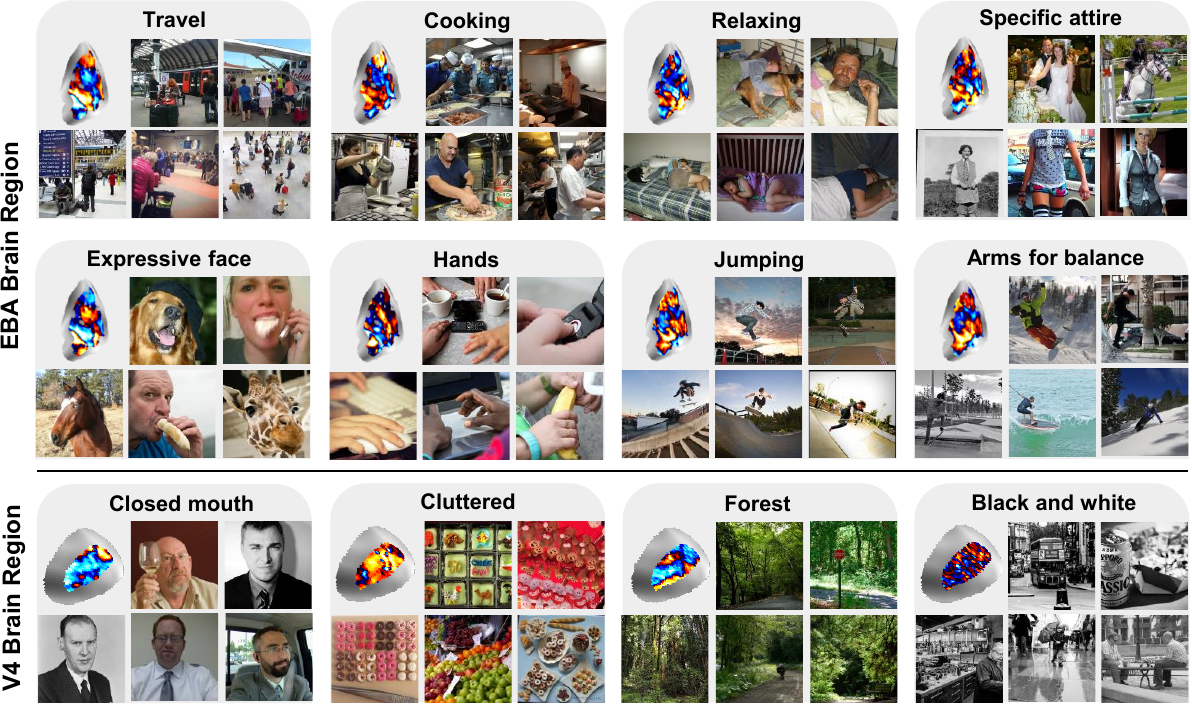}
  \caption{\textbf{Discovered Interpretable Patterns of Subject 5 (EBA and hV4)}. We show patterns for subject 5 with top activating images and selected explanations. EBA is known to encode bodies and actions and V4 is known to encode mid-level features (e.g., color, shape).}  
  \label{fig:sub5_concepts_2}
\end{figure*}

\vspace{-0.15cm}
\paragraph{Additional subjects visualization.}
We present discovered interpretable patterns and concepts for two additional subjects with a larger number of examples (Subjects~2 and~5) in \cref{fig:sub2_concepts_1,fig:sub2_concepts_2,fig:sub5_concepts_1,fig:sub5_concepts_2}. All subjects’ trained decompositions and discovered patterns will be publicly available, enabling further research and discoveries.

\begin{figure*}[ht]
  \centering
\includegraphics[width=\linewidth]{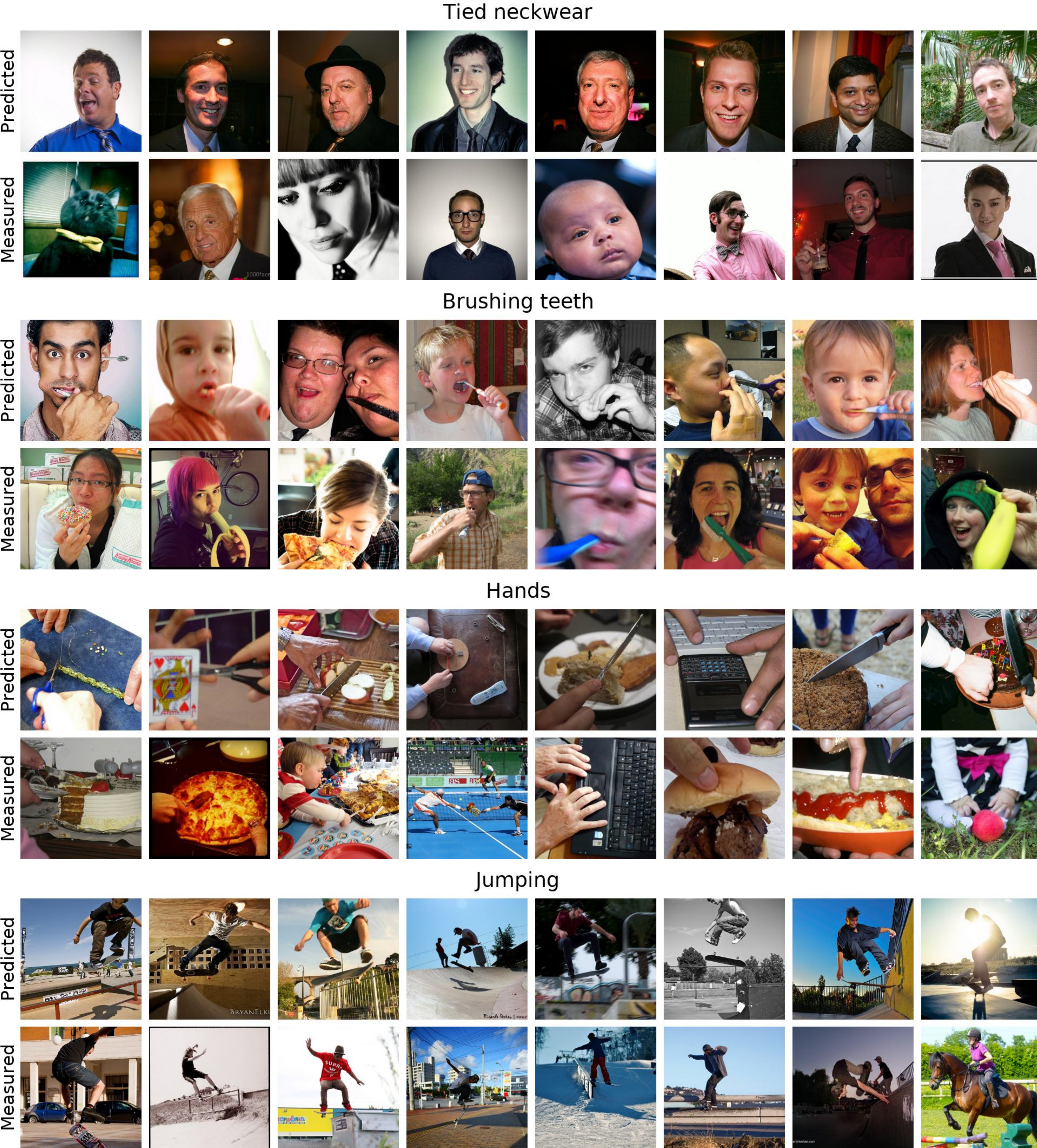}
  \caption{{\textbf{Image grid EBA (Part 1).} Full visual grid with top 16 activating images for concepts \textit{Tied neckwear}, \textit{Brushing teeth}, \textit{Hands}, and \textit{Jumping} found in the EBA region. For each concept, the top 8 images corresponding to \textbf{Measured} and \textbf{Predicted} fMRI responses are shown.}}
  \label{fig:eba_grid_1}
\end{figure*}

\begin{figure*}[ht]
  \centering
\includegraphics[width=\linewidth]{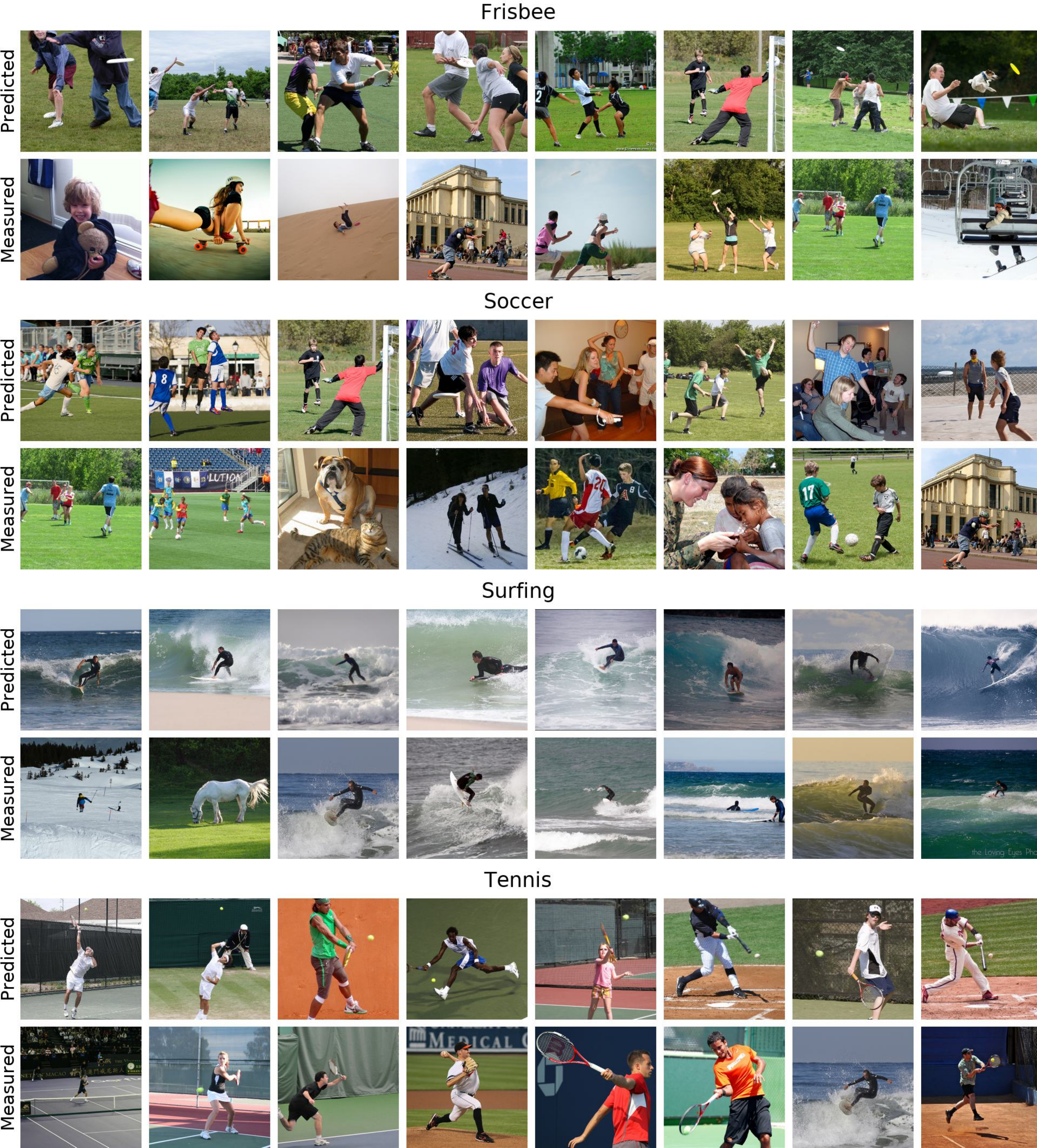}
  \caption{\textbf{Image grid EBA (Part 2).} Full visual grid with top 16 activating images for concepts \textit{Frisbee}, \textit{Soccer}, \textit{Surfing}, and \textit{Tennis} found in the EBA region. For each concept, the top 8 images corresponding to \textbf{Measured} and \textbf{Predicted} fMRI responses are shown.}
  \label{fig:eba_grid_2}
\end{figure*}

\begin{figure*}[ht]
  \centering
\includegraphics[width=\linewidth]{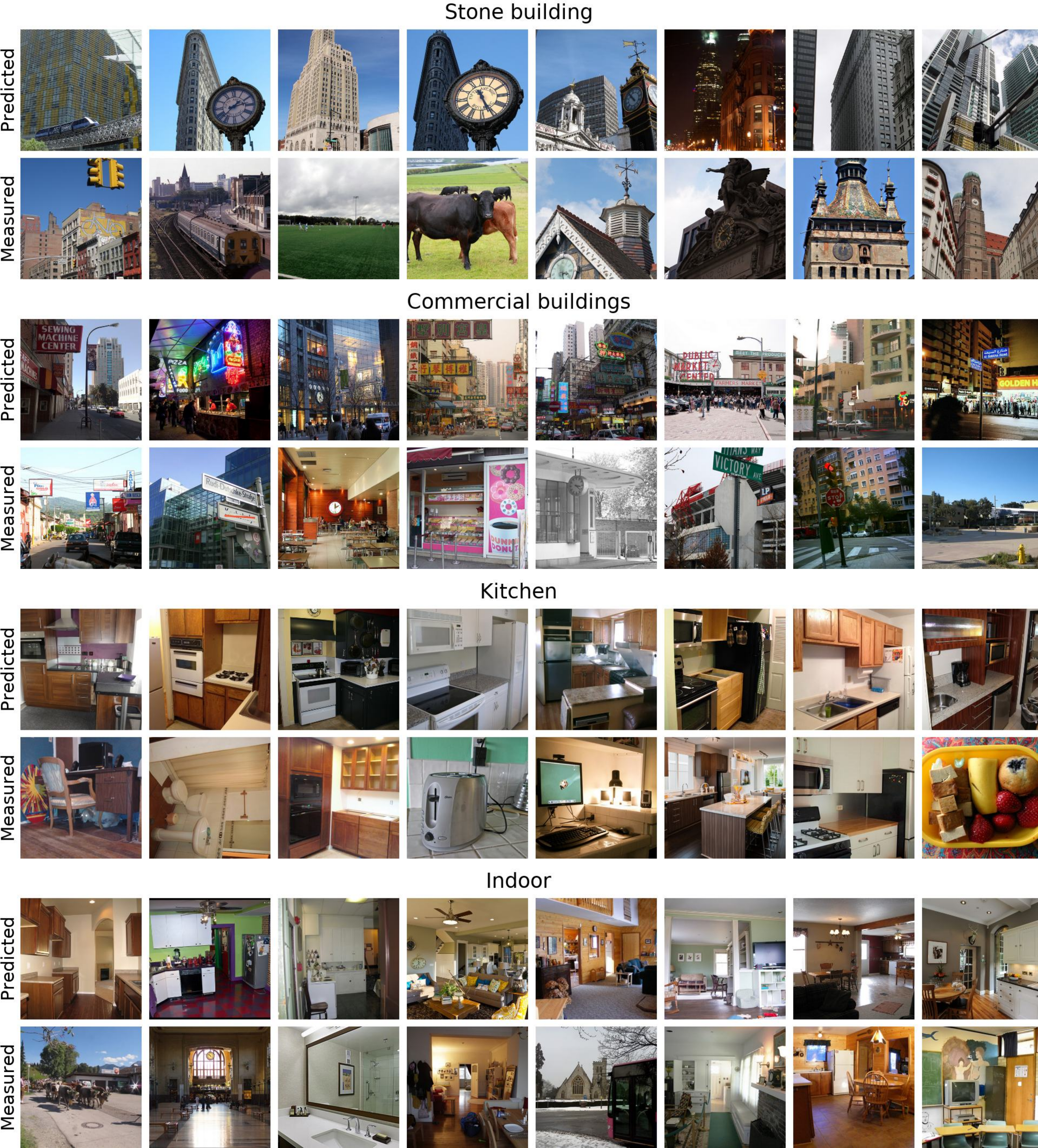}
  \caption{\textbf{Image grid PPA (Part 1).} Full visual grid with top 16 activating images for concepts \textit{Stone building}, \textit{Commercial buildings}, \textit{Kitchen}, and \textit{Indoor} found in the PPA region. For each concept, the top 8 images corresponding to \textbf{Measured} and \textbf{Predicted} fMRI responses are shown.}
  \label{fig:ppa_grid_1}
\end{figure*}

\begin{figure*}[ht]
  \centering
\includegraphics[width=\linewidth]{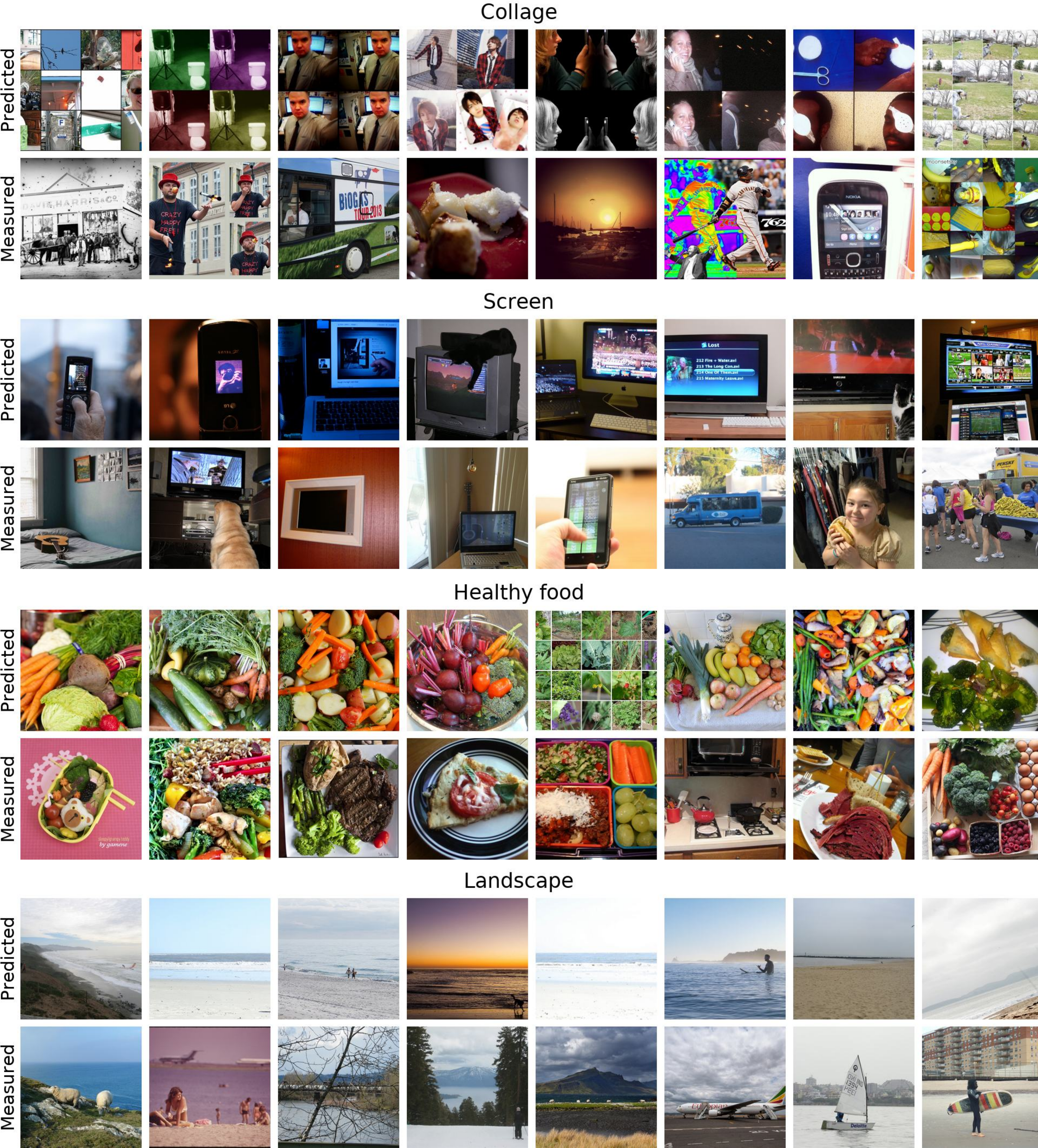}
  \caption{\textbf{Image grid PPA (Part 2).} Full visual grid with top 16 activating images for concepts \textit{Collage}, \textit{Screen}, \textit{Healthy food}, and \textit{Landscape} found in the PPA region. For each concept, the top 8 images corresponding to \textbf{Measured} and \textbf{Predicted} fMRI responses are shown.}
  \label{fig:ppa_grid_2}
\end{figure*}

\vspace{-0.15cm}
\paragraph{Full top 16 images per pattern.}
We present full visual grids showing the top 16 activating images for the concepts from EBA discussed in the main paper (\cref{fig:eba_grid_1,fig:eba_grid_2}). For each concept, the top row corresponds to predicted fMRI responses and the bottom row to measured fMRI responses. Images are ordered by activation strength, from left to right. We also show all 16 top-activating images for the patterns and concepts presented in the main paper for PPA (\cref{fig:ppa_grid_1,fig:ppa_grid_2}).

\begin{figure*}[ht]
  \centering
\includegraphics[width=\linewidth]{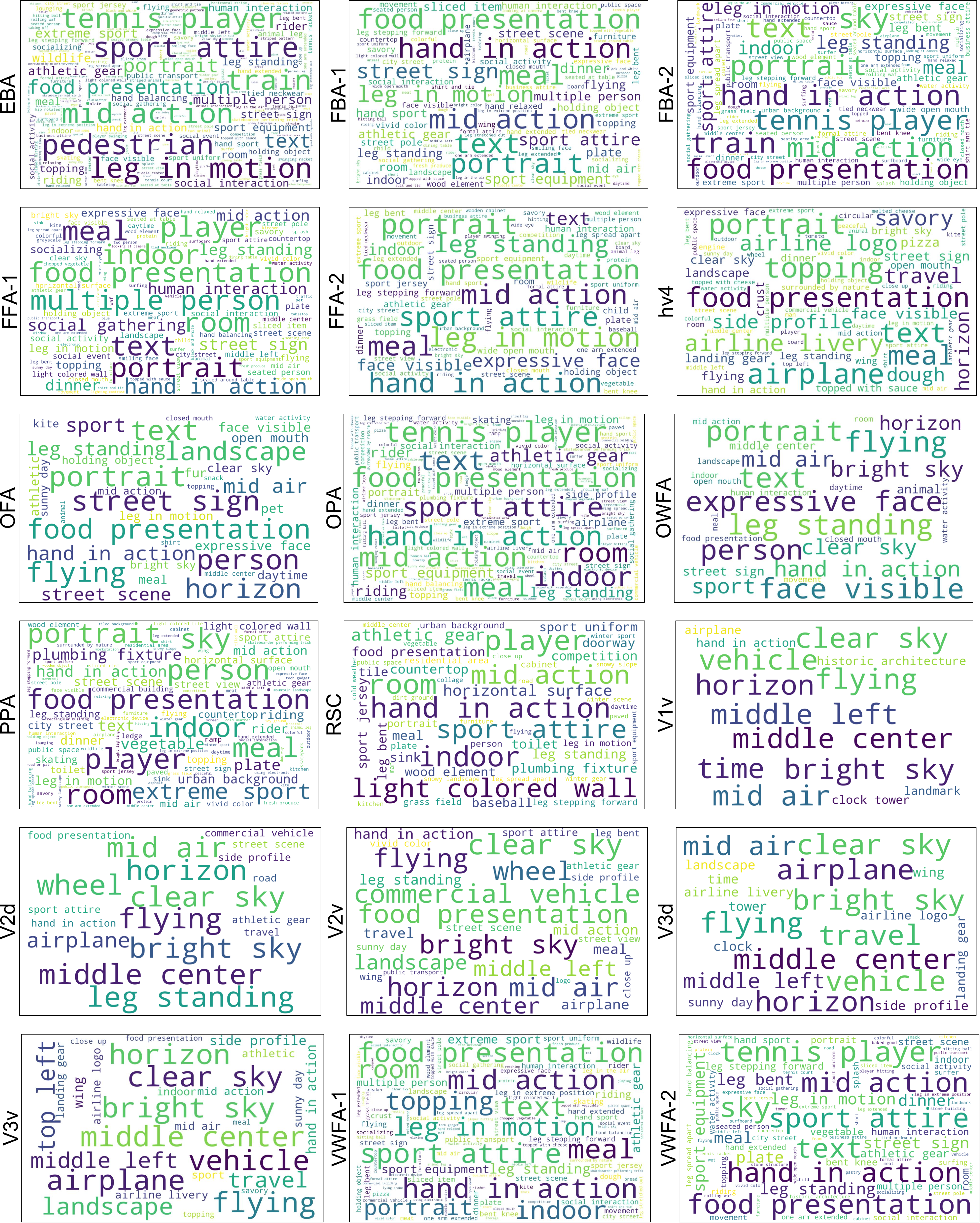}
  \vspace{-0.62cm}
  \caption{\textbf{Concepts best explained by every ROI (Non-Exlusive)}. Concepts represented in each Region of Interest (ROI) that achieve an alignment score $> 0.5$. A concept may appear in multiple ROI word clouds, reflecting its representation across different brain regions. Word size reflects the alignment score of the concept within the assigned ROI.}
  \vspace{-0.6cm}
  \label{fig:word_cloud}
\end{figure*}

\begin{figure*}[ht]
  \centering
\includegraphics[width=\linewidth]{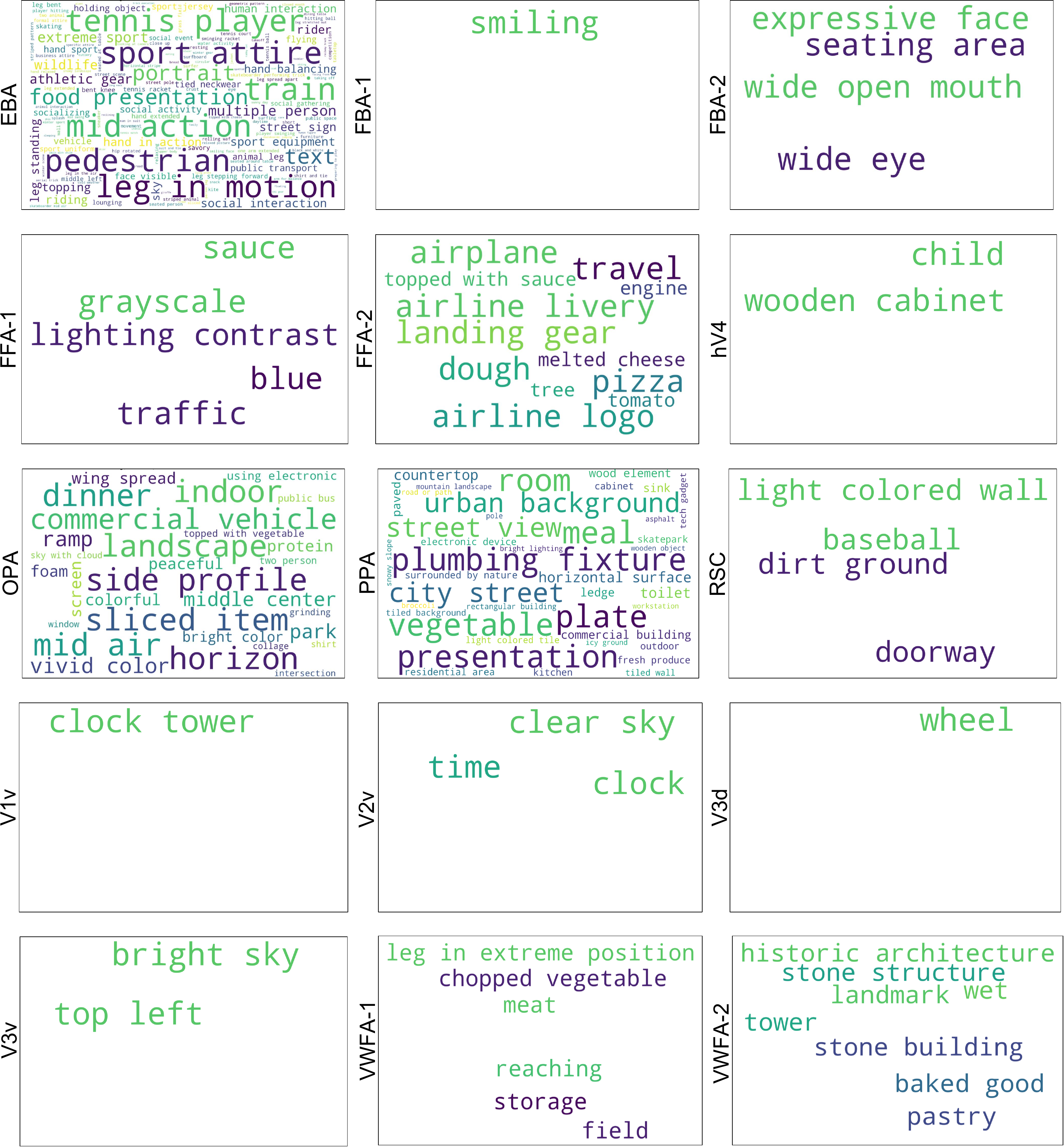}
  \caption{\textbf{Concepts best explained by every ROI (Exlusive)}. Each concept is assigned to a single ROI—the one with the highest alignment score. Only concepts with alignment $> 0.5$ are shown. Word size reflects the alignment score within the assigned ROI.}  
  \label{fig:word_cloud_excluded}
\end{figure*}

\vspace{-0.15cm}
\paragraph{Concepts best explained by each ROI.}
We summarize which semantic concepts are most strongly represented across different Regions of Interest (ROIs). \cref{fig:word_cloud} shows, for each ROI, the concepts achieving an alignment score above 0.5, allowing a concept to appear in multiple ROIs if it is represented across regions. In contrast, \cref{fig:word_cloud_excluded} presents an exclusive version in which each concept is assigned only to the ROI where it achieves its highest alignment score. Together, these visualizations reveal both the distributed and the region-specific organization of conceptual representations across the visual cortex.

\begin{figure*}[ht]
  \centering
\includegraphics[width=\linewidth]{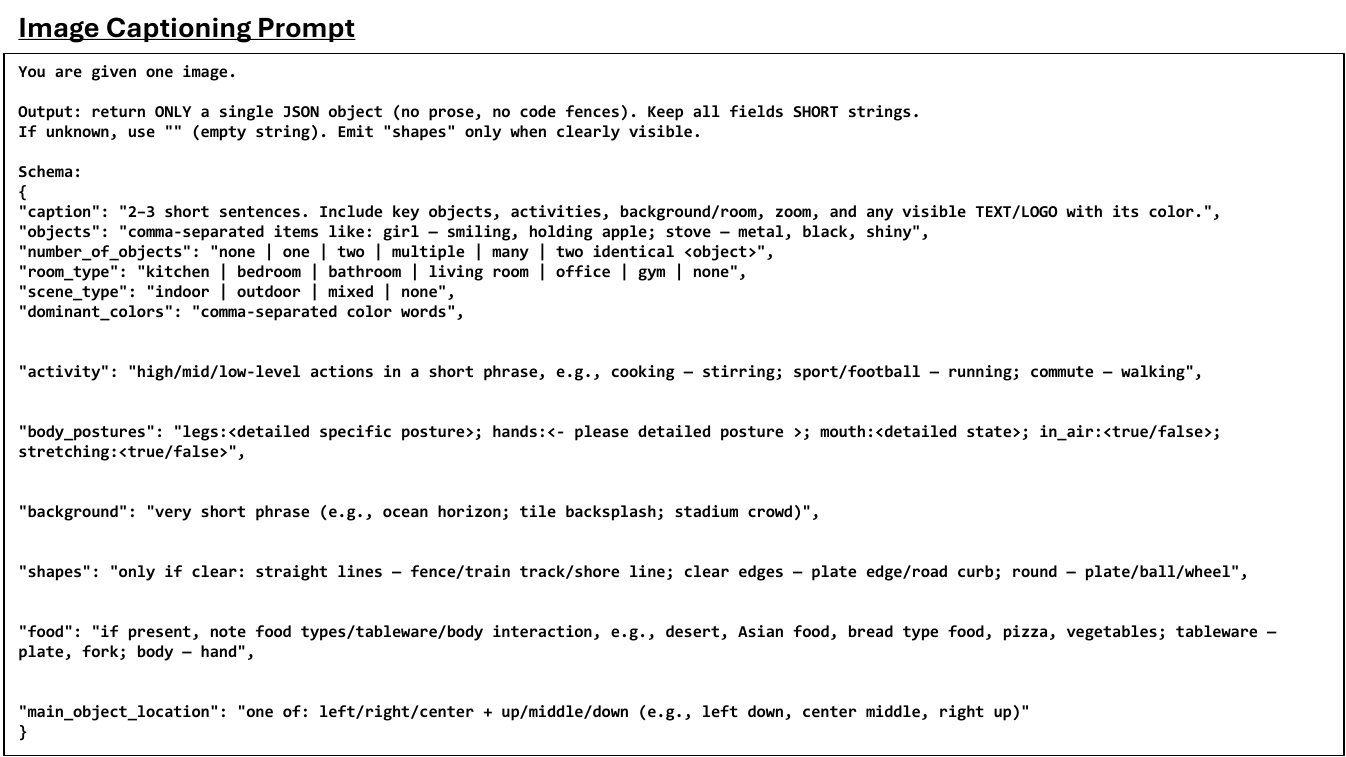}
\caption{\textbf{Image Captioning Prompt.}}
  \label{sup_fig:image_captioning_prompt}
\end{figure*}

\begin{figure*}[ht]
  \centering
\includegraphics[width=\linewidth]{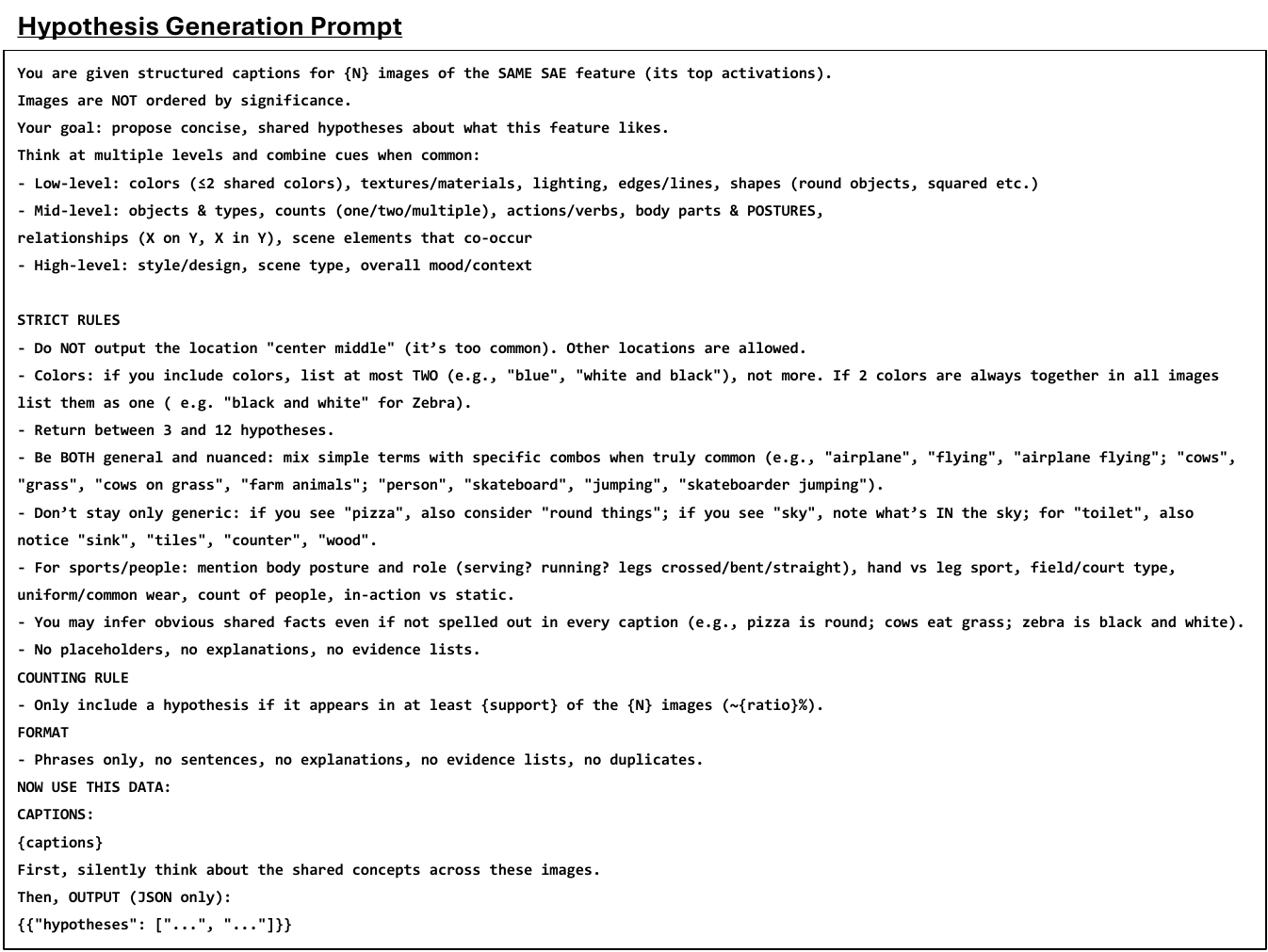}
\caption{\textbf{Hypothesis Generation Prompt.}
Given a set of 10 images and their detailed captions, an LLM is instructed to identify what is common across the images and to generate hypotheses that may explain what is shared among them.}
  \label{sup_fig:Hyp_gen_prompt}
\end{figure*}

\begin{figure*}[ht]
  \centering
\includegraphics[width=\linewidth]{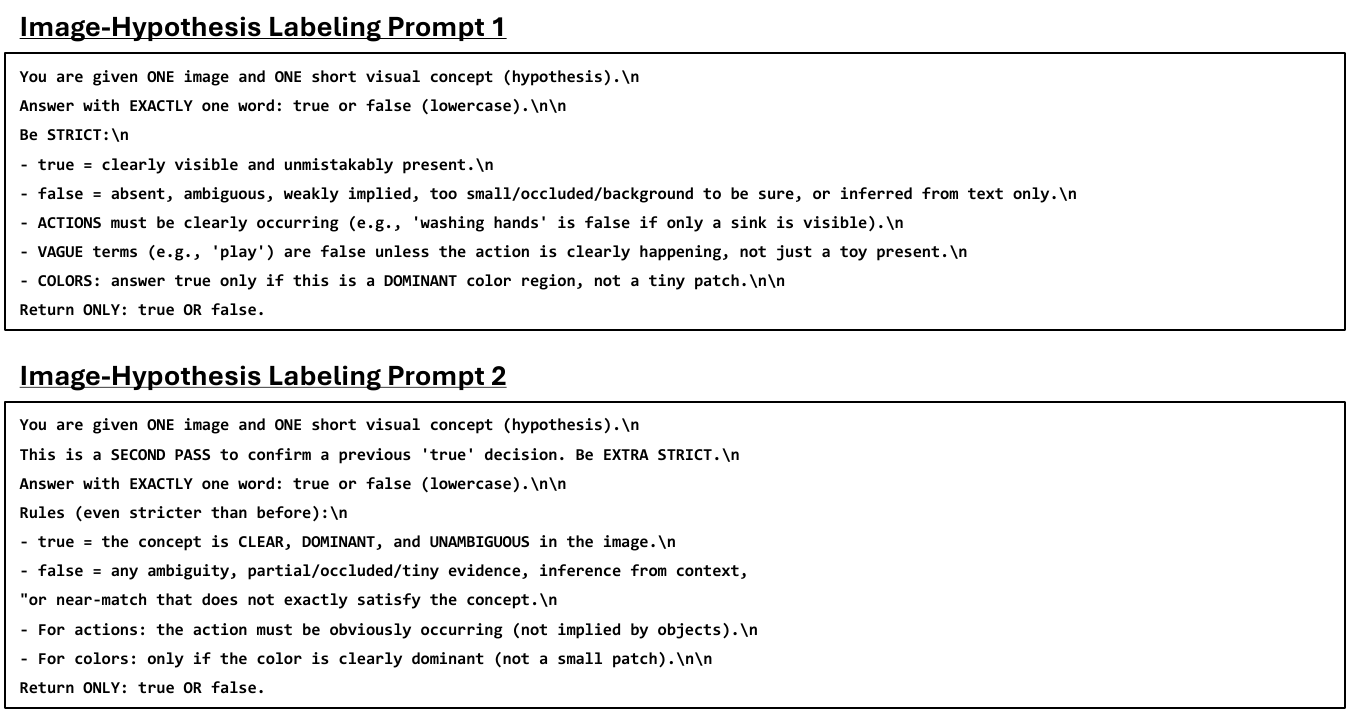}
\caption{\textbf{Image–Hypothesis Labeling Prompt.}
For each image–hypothesis pair, we label whether the hypothesis is supported by the image or not.}
  \label{sup_fig:Hyp_label_prompt}
\end{figure*}

\vspace{-0.15cm}
\paragraph{Prompts.}
\Cref{sup_fig:image_captioning_prompt,sup_fig:Hyp_gen_prompt,sup_fig:Hyp_label_prompt} depict the prompts used in our pipeline: first, detailed per-image captions are elicited (\cref{sup_fig:image_captioning_prompt}); second, a language model infers shared hypotheses across the image set (\cref{sup_fig:Hyp_gen_prompt}); and, distinct from the two-stage explanation, the image–hypothesis labeling prompt is used in the \emph{upscale} stage to decide whether each hypothesis is supported by each image (\cref{sup_fig:Hyp_label_prompt}).

\begin{figure*}[ht]
  \centering
\includegraphics[width=\linewidth]{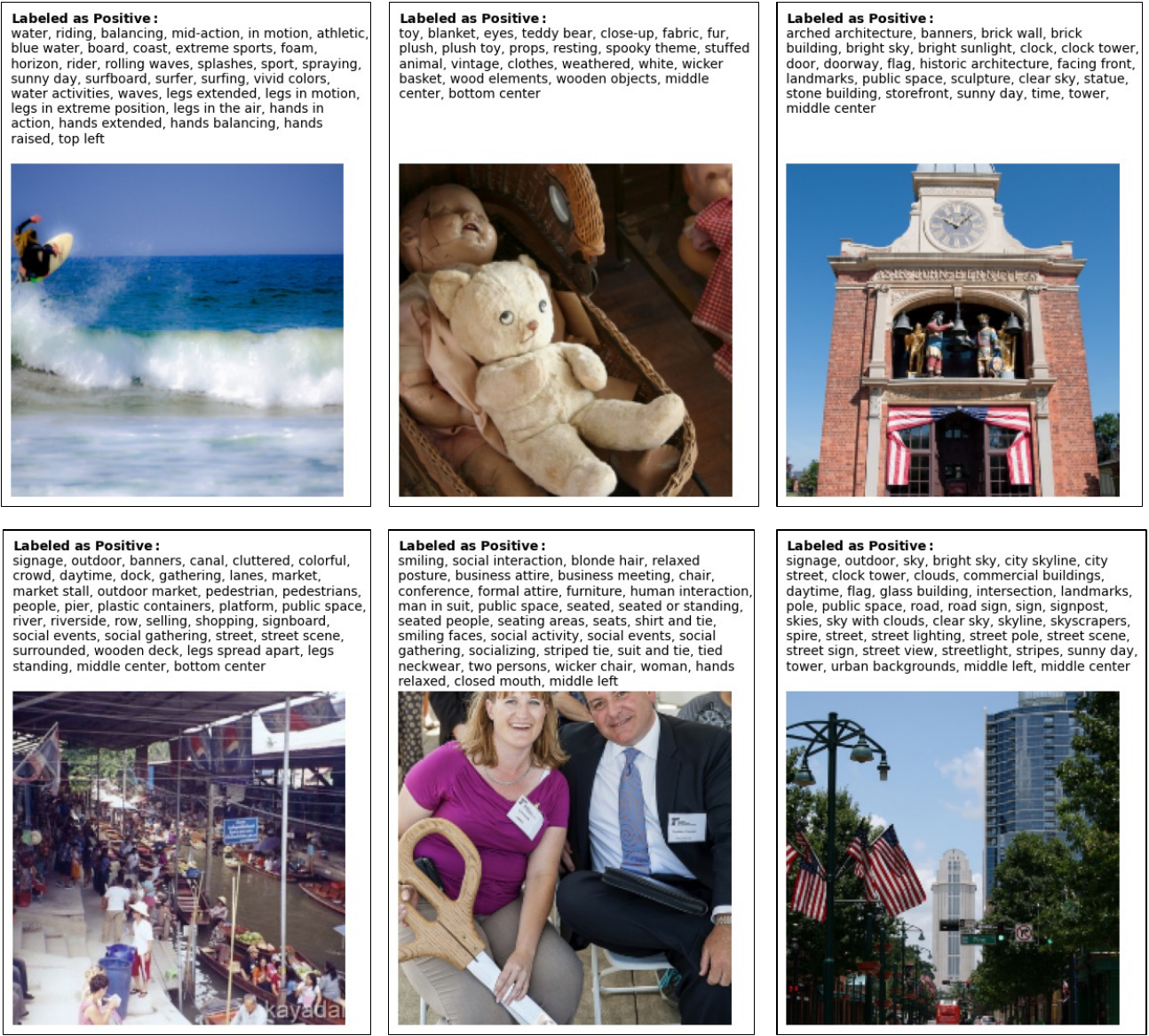}
\caption{\textbf{Images with per-hypothesis labeling.}
We show six example images and list all hypotheses labeled as positive for each.
The hypotheses are taken from our brain-inspired hypothesis dictionary, and the images are drawn from the set used with predicted fMRI.}
  \label{sup_fig:image_labeled}
\end{figure*}

\vspace{-0.15cm}
\paragraph{Labeled examples.}
\Cref{sup_fig:image_labeled} presents representative images together with the hypotheses labeled as positive for each, demonstrating the output produced by the hypothesis–image labeling stage.

\end{document}